\documentclass[journal]{IEEEtran}
\usepackage{times}  
\usepackage{helvet}  
\usepackage{courier}  
\usepackage{url}  
\usepackage{graphicx}  
\usepackage{amsfonts}
\usepackage{bm}
\usepackage{multirow}
\graphicspath{{figures/}{figures/tem/}}
\usepackage{epstopdf}
\usepackage{cite}
\usepackage{amsmath}
\usepackage{hyperref}
\usepackage{booktabs}
\usepackage{setspace}
\usepackage{amsmath}
\usepackage{array}
\usepackage{color}
\usepackage{algorithm}
\usepackage{algorithmic}
\usepackage{subcaption}  
\usepackage{float}
\usepackage{xcolor}  
\usepackage{pifont}  
\usepackage{multirow}
\usepackage{makecell}
\newtheorem{theorem}{Theorem} 
\newtheorem{definition}{Definition}

\ifCLASSINFOpdf
\else
\fi
\begin{document}
%
\title{Stability Enhancement in Reinforcement Learning via Adaptive Control Lyapunov Function}
%
%
%

\author{Donghe~Chen, Han~Wang, Lin~Cheng, Shengping~Gong

  \thanks{Authors' address: Donghe Chen, Han Wang, Lin~Cheng and Shengping Gong  are with the School of Astronautics, Beihang University, 102206 Beijing, China  (Corresponding author: Lin Cheng, chenglin5580@buaa.edu.cn.)}


}

%
%

\markboth{Journal of IEEE Transactions on Aerospace and Electronic Systems}
{Donghe, Chen.;Han, Wang.;Lin, Cheng.; and Shengping, Gong.: Stability Enhancement in Reinforcement Learning via Adaptive Control Lyapunov Function}
%



\maketitle

\begin{abstract}
  Reinforcement Learning (RL) has shown promise in control tasks but faces significant challenges in real-world applications, primarily due to the absence of safety guarantees during the learning process. Existing methods often struggle with ensuring safe exploration, leading to potential system failures and restricting applications primarily to simulated environments. Traditional approaches such as reward shaping and constrained policy optimization can fail to guarantee safety during initial learning stages, while model-based methods using Control Lyapunov Functions (CLFs) or Control Barrier Functions (CBFs) may hinder efficient exploration and performance. To address these limitations, this paper introduces Soft Actor-Critic with Control Lyapunov Function (SAC-CLF), a framework that enhances stability and safety through three key innovations: (1) a task-specific CLF design method for safe and optimal performance; (2) dynamic adjustment of constraints to maintain robustness under unmodeled dynamics; and (3) improved control input smoothness while ensuring safety. Experimental results on a classical nonlinear system and satellite attitude control demonstrate the effectiveness of SAC-CLF in overcoming the shortcomings of existing methods.
\end{abstract}


%

\section{Introduction}
\label{sec:intro}

\IEEEPARstart{I}{n} aerospace engineering, classic control methods like PID remain widely used due to the significant effort required to develop cost-effective and efficient dynamic models for complex systems\cite{zhao_-board_2023, li_closed-loop_2023, wu_low-thrust_2024, wang_spacecraft_2024}. This challenge has led the control community to explore schemes that achieve satisfactory performance with less precise model knowledge.  Robust control\cite{ionescu_robust_2020} exemplifies this approach by accommodating modeling inaccuracies while ensuring a minimum performance level. Adaptive control\cite{qu_dynamic-matching_2024,wang_adaptive_2017}, on the other hand, enables controllers to adjust their parameters in real-time for optimal performance, further reducing reliance on detailed system models.

Among these approaches, Reinforcement Learning (RL) stands out as a model-free method where agents learn optimal behaviors through trial and error, guided by rewards or penalties. RL focuses on training an agent to make decisions that maximize long-term rewards. Through iterative cycles of observation, action selection, execution, and feedback, the agent continuously refines its strategy to optimize cumulative rewards \cite{lillicrap_continuous_2019, schulman_trust_2017}. This learning process has shown significant promise, particularly in continuous control applications, leading to advanced developments in robotics after extensive training \cite{zhao_sim--real_2020, ohnishi_safety-aware_2018}.

In model-free reinforcement learning (RL), the agent learns to maximize long-term rewards through an exploratory phase, where it tries new actions to discover their outcomes \cite{shah_airsim_2017, ames_control_2017}. This exploration is crucial for learning but can lead to potentially unsafe behaviors, posing significant risks when applied to real-world hardware \cite{berkenkamp_safe_2017}. Balancing exploration and exploitation is critical: transitioning too early to exploitation can result in suboptimal behavior due to insufficient knowledge of the environment, while delaying this transition unnecessarily can impact performance.

Given these challenges, many successful RL applications in physical system control have remained within simulated environments, where the trial-and-error process does not carry the same level of risk \cite{mnih_human-level_2015, mnih_playing_2013, silver_mastering_2017}. The trade-off between exploring new actions and exploiting known rewarding behaviors—known as the exploration-exploitation dilemma—remains a key challenge for deploying RL algorithms in real-world systems. In these controlled simulations, researchers can safely experiment with different strategies, refining algorithms before considering their application in more hazardous real-world scenarios.

However, transitioning from simulation to real-world applications introduces additional complexities. Switching between primary and backup controllers designed for safety can overly restrict the exploration of potential policies \cite{nguyen-tuong_local_2008,yang_safe_2022}. When an agent attempts inappropriate actions, it receives penalties that progressively discourage such behavior. In hazardous environments, erroneous actions can lead to irreversible long-term consequences, making the balance between safety and exploration particularly challenging. Thus, the objective is to prevent these issues during the learning process to ensure safe exploration \cite{pecka_safe_2014}.  To achieve this, integrating safety mechanisms into the learning framework becomes essential, allowing for cautious yet effective exploration in real-world settings without compromising system integrity.

Safe Reinforcement Learning (SRL) optimizes returns under safety constraints, addressing limitations of traditional methods such as reward shaping and constrained policy optimization, which often fail to ensure safety during initial learning stages \cite{gu_review_2024}. Model-based approaches, including Control Lyapunov Functions (CLFs) or Control Barrier Functions (CBFs) and model predictive control, provide safety assurances under known dynamics but may restrict efficient exploration and performance \cite{berkenkamp_safe_2017, wang_deepsafempc_2024}. CLF-based mechanisms face challenges: lack of automated design methods for complex nonlinear systems necessitating domain expertise \cite{ames_control_2017}, inadaptability to unmodeled dynamics due to static parameters undermining robustness \cite{fisac_general_2019}, and neglect of control input smoothness leading to abrupt changes that can degrade system performance. These limitations emphasize the need for advanced SRL techniques to handle dynamic and uncertain environments effectively.

Therefore, developing advanced SRL techniques that overcome these limitations is crucial for achieving both safe and effective exploration in real-world applications. By addressing these challenges, researchers aim to bridge the gap between theoretical models and practical implementation, ensuring that RL can be reliably deployed in safety-critical environments. To address the challenges of designing adaptive and safe control policies in dynamic environments, we introduce SAC-CLF, an integration of Control Lyapunov Functions (CLFs) within the Soft Actor-Critic (SAC) framework \cite{haarnoja_soft_2018}. This approach enhances policy exploration by ensuring adherence to safety constraints and optimizing performance. The quadratic programming (QP) form of the controllers has proven effective in real-time applications, such as bipedal walking on human-sized robots \cite{galloway_torque_2015} and controlling scale cars \cite{mehra_adaptive_2015}. By incorporating a CLF-based safety mechanism, SAC-CLF leverages the stability and sample efficiency of the model-free SAC algorithm while actively guiding policy updates to comply with predefined safety requirements.

Our contributions are summarized as follows:

\begin{enumerate}
  \item \textbf{Task-Specific CLFs Design}: We introduce a method for designing quadratic CLFs via system linearization and LQR techniques, addressing the challenge of lacking systematic approaches for complex nonlinear systems. This ensures learned policies are safe and accelerates learning by focusing exploration on safety-compliant strategies.

  \item \textbf{Adaptive Constraint Strength}: We propose an adaptive adjustment method for constraint strength based on discrepancies between desired and actual CLF derivatives. This enhances robustness and optimality in the presence of unknown dynamics or disturbances.
  
  \item \textbf{Safety-Prioritized Control Input Smoothing}: A vibration-dampening term is added to the objective function, ensuring smooth control inputs while maintaining safety, thus improving operational efficiency and reducing actuator wear.
\end{enumerate}

\section{Preliminaries}
This section introduces the problems and methodologies in nonlinear control theory for continuous-time systems. First, the system dynamics and control policy are described.Second, Soft Actor-Critic (SAC) in continuous time is analyzed. Finally, Control Lyapunov Functions (CLFs) are examined for ensuring stability, while challenges in CLF design.

\subsection{Problem Statement}
An infinite-horizon discounted Markov Decision Process (MDP) with control-affine dynamics is considered. The state transition in a continuous-time setting is modeled using ordinary differential equations. Let $\boldsymbol{x}(t) \in \mathcal{X}(\mathcal{X} \subseteq \mathbb{R}^n)$ denote the state of the system at time $t$, and let $\boldsymbol{u}(t) \in \mathcal{U} (\mathcal{U} \subseteq  \mathbb{R}^m)$ represent the control input applied at time $t$. The dynamics of the system are described by:
\begin{equation}
  \label{eq:dynamic_model}
  \begin{aligned}
    \dot{\boldsymbol{x}}(t) &= \boldsymbol{f}(\boldsymbol{x}(t)) + \boldsymbol{g}(\boldsymbol{x}(t))\boldsymbol{u}(t) + \boldsymbol{d}(\boldsymbol{x}(t))\\
    \boldsymbol{e}(t) &= \boldsymbol{x}(t) - \boldsymbol{x}_d(t)
  \end{aligned}
\end{equation}
where $\dot{\boldsymbol{x}}(t)$ denotes the time derivative of the state, $\boldsymbol{f}: \mathbb{R}^n \rightarrow \mathbb{R}^n$ represents the drift term corresponding to unforced system evolution, $\boldsymbol{g}: \mathbb{R}^n \rightarrow \mathbb{R}^{n \times m}$ serves as the input gain matrix, and $\boldsymbol{d}: \mathbb{R}^n \rightarrow \mathbb{R}^n$ captures the unknown or uncertain system dynamics. The functions $\boldsymbol{f}$ and $\boldsymbol{g}$ are assumed to be known.

To enhance controller design, we define the extended state vector $\boldsymbol{x}_e(t) = [\boldsymbol{x}(t)^\top, \boldsymbol{e}(t)^\top]^\top$, which integrates the system state $\boldsymbol{x}(t)$—capturing operational variables—and the error state $\boldsymbol{e}(t)$—representing deviation from a desired state. The objective is to find an optimal control policy $\pi(\boldsymbol{u}|\boldsymbol{x})$ that minimizes the expected infinite-horizon discounted cost:

\begin{equation}
  J(\pi) = \mathbb{E}_{\pi} \left\{ \int_{0}^{\infty} e^{-\gamma t} \cdot r(\boldsymbol{x}_e(t), \boldsymbol{u}(t)) dt \right\}
\end{equation}

The instantaneous cost function is defined as $r(\boldsymbol{x}_e(t), \boldsymbol{u}(t)) = r_e(\boldsymbol{x}_e(t)) + \boldsymbol{u}(t)^\top \boldsymbol{R} \boldsymbol{u}(t)$, where $r_e(\boldsymbol{x}_e(t))$ penalizes deviations from the desired state and $\boldsymbol{u}(t)^\top \boldsymbol{R} \boldsymbol{u}(t)$ imposes a penalty on control effort, with $\boldsymbol{R}$ being a positive definite weighting matrix. The discount factor $\gamma > 0$ prioritizes immediate costs over future ones in the cost evaluation.

In summary, the aim is to determine the policy $\pi^*$ that achieves the lowest $J(\pi)$, effectively balancing state regulation and control effort while considering both immediate and future costs.

\subsection{Soft Actor-Critic }

The Soft Actor-Critic (SAC) algorithm \cite{haarnoja_soft_2018} aims to find an optimal policy $\pi^*$ that maximizes an objective function which integrates both the expected accumulated discounted reward and the entropy of the policy. This dual-objective promotes both high performance and exploratory behavior, leading to more robust policies. The objective function is defined as:

\begin{equation}
    J(\pi) = \mathbb{E}_{\pi} \left\{ \sum_{t=0}^{\infty} \gamma^t \left[ r(\boldsymbol{x}_{e}(t),  \boldsymbol{u}(t)) + \alpha H(\pi(\cdot|\boldsymbol{x}_{e}(t))) \right] \right\}
\end{equation}

where $r: \mathcal{X} \times \mathcal{X} \times \mathcal{U} \rightarrow \mathbb{R}$ represents the scalar cost associated with each state-input pair $(\boldsymbol{x}_{e}(t), \boldsymbol{u}(t))$, $\gamma \in (0, 1)$ is the discount factor ensuring convergence of the infinite sum, and $\alpha < 0$ is a hyperparameter that balances the trade-off between maximizing expected reward and promoting exploration through the entropy term $H(\pi(\cdot|\boldsymbol{x}_{e}))$.

In this framework, the entropy-regularized state-input value function $Q_H$ plays a crucial role and is defined as:

\begin{equation}
    Q_H(\boldsymbol{x}_{e}, \boldsymbol{u}) = Q(\boldsymbol{x}_{e}, \boldsymbol{u}) + \alpha H(\pi(\cdot|\boldsymbol{x}_{e}))
\end{equation}

The optimal policy $\pi^*$ is derived from optimizing $Q_H$. Specifically, for each state-input pair, the optimal input minimizes the following expression:

\begin{equation}
    Q^*(\boldsymbol{x}_{e}, \boldsymbol{u}) = r(\boldsymbol{x}_{e}, \boldsymbol{u}) + \gamma \mathbb{E}_{\boldsymbol{x}_{e}' \sim p(\cdot|\boldsymbol{x}_{e}, \boldsymbol{u})} \left[ \min_{\boldsymbol{u}'} Q_H(\boldsymbol{x}_{e}', \boldsymbol{u}') \right]
\end{equation}
where $\boldsymbol{x}_{e}'$ denotes the next state obtained after applying control $\boldsymbol{u}$ at state $\boldsymbol{x}_{e}$, and $p(\cdot|\boldsymbol{x}_{e}, \boldsymbol{u})$ is the transition probability from state $\boldsymbol{x}_{e}$ to $\boldsymbol{x}_{e}'$ given input $\boldsymbol{u}$.

\subsection{Integrating Control Lyapunov Functions with Quadratic Programming}

CLFs provide a sufficient condition for ensuring global asymptotic stability. A positive definite, continuously differentiable function $V: \mathbb{R}^n \to \mathbb{R}$ is defined as the potential energy or "cost-to-go" of the system state. For global asymptotic stability to be ensured, the convergence of the system's state to the zero-error over time must occur.

The existence of a CLF indicates that a controller $\boldsymbol{u}^{\text{CLF}}$ can be designed to ensure the closed-loop system's global asymptotic stability. The controller aims to satisfy the CLF condition while minimizing control effort, as given by:
\begin{equation}
  \begin{aligned}
    (\boldsymbol{u}(t), \varepsilon) &= \mathop{\arg\min}\limits_{\boldsymbol{u} \in \mathbb{R}^{m}, \varepsilon \in \mathbb{R}^{+}} \|\boldsymbol{u}(t) - \boldsymbol{u}_{ref}(t)\|_2^2 + K_\varepsilon \varepsilon \\
    \text{s.t.} \quad &\frac{\partial V}{\partial \boldsymbol{x}} [\boldsymbol{f}(\boldsymbol{x}(t)) + \boldsymbol{g}(\boldsymbol{x}(t))\boldsymbol{u}(t)] + \eta V(\boldsymbol{e}(t)) \leq \varepsilon \\
    & \boldsymbol{u}_{\text{low}}^{i} \leq \boldsymbol{u}^{i}(t) \leq \boldsymbol{u}_{\text{high}}^{i}, \quad i=1,\ldots,m
  \end{aligned}
  \label{eq:QP_raw}
\end{equation}
where $\eta > 0$ controls the convergence rate; $\boldsymbol{u}_{\text{low}}$, $\boldsymbol{u}_{\text{high}}$ are the actuator limits; $\varepsilon$ is a slack variable for flexibility; and $K_\varepsilon$ is a large positive penalty coefficient enforcing strict constraints. The objective minimizes the 2-norm of the deviation between the control input $\boldsymbol{u}(t)$ and a reference input $\boldsymbol{u}_{ref}(t)$, along with the slack variable, promoting exponential stability. 

\section{CLF-Based Compensating Control with Reinforcement Learning}

\begin{figure}[htbp]
  \centering
  \includegraphics[width=0.45\textwidth]{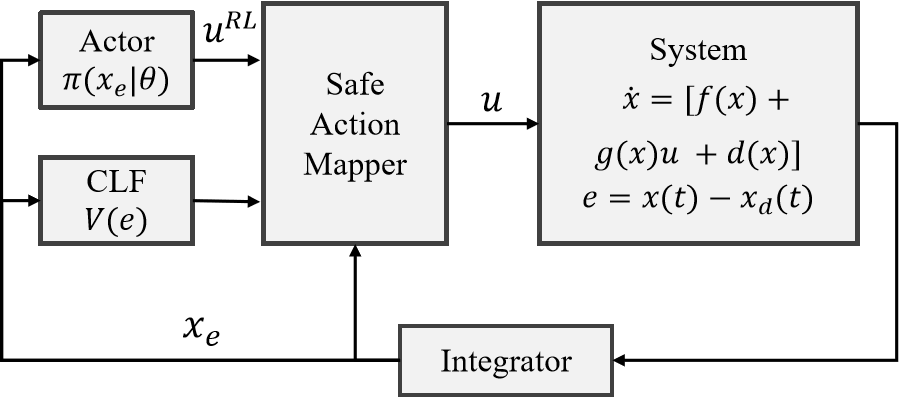}
  \caption{Framework of SAC-CLF}
  \label{fig:framework}
\end{figure}

The SAC-CLF framework, illustrated in Fig.~\ref{fig:framework} and utilizing SAC as its foundation, is significantly enhanced through three key improvements: (1) To address the challenge of lacking a systematic  approach for CLF design, especially for complex and nonlinear systems, a method for designing a quadratic Control CLF via system linearization and Linear Quadratic Regulator (LQR) techniques is introduced. (2) To tackle the problem of inadaptability to disturbances and enhance the robustness of the control policy in dynamic and uncertain environments, the framework adjusts the conservatism of constraints based on the discrepancy between the desired and actual CLF derivatives. (3) To mitigate abrupt changes in control inputs that can cause wear and tear on hardware and degrade system performance, a vibration-dampening term is incorporated into the objective function.

\subsection{Task-Specific CLFs Design}

In control theory, designing a CLF that is well-matched to the task at hand is crucial for ensuring system stability while guiding the system according to predefined mission requirements \cite{ames_rapidly_2014}. A properly designed CLF not only guarantees asymptotic stability but also directs the system's evolution in line with desired objectives. To achieve this, we first define the general properties of a CLF and introduce the concept of the nominal system, then proceed to discuss Safe State Space and Safe Energy Ball, and finally prove that if the initial state lies within the safe energy ball, all subsequent states will remain within this region.

\begin{definition}[Nominal System]
  The nominal system is an idealized model of a dynamic system operating under standard conditions, without external disturbances. Its behavior is described by:

  \begin{equation}
    \label{eq:dynamic_model}
    \begin{aligned}
      \dot{\boldsymbol{x}}^*(t) &= \boldsymbol{f}(\boldsymbol{x}^*(t)) + \boldsymbol{g}(\boldsymbol{x}^*(t))\boldsymbol{u}(t),\\
      \boldsymbol{e}(t) &= \boldsymbol{x}^*(t) - \boldsymbol{x}^*_d(t),
    \end{aligned}
  \end{equation}

  where $\dot{\boldsymbol{x}^*}(t)$ is the state derivative, $\boldsymbol{f}(\cdot)$ and $\boldsymbol{g}(\cdot)$ represent the system dynamics, $\boldsymbol{u}(t)$ is the control input, and $\boldsymbol{e}(t)$ is the error relative to the desired state $\boldsymbol{x}^*_d(t)$.
\end{definition}

\begin{theorem}[Control Lyapunov Function]
  A function $V: \mathbb{R}^n \to \mathbb{R}$ is termed a control Lyapunov function for a system if it satisfies:

  \begin{enumerate}
      \item \textbf{Continuously Differentiable:} $V$ is continuously differentiable on $\mathbb{R}^n$.
      
      \item \textbf{Positive Definite:} $V(0) = 0$ and $V(\boldsymbol{e}) > 0$ for all $\boldsymbol{e} \neq 0$.
      
      \item \textbf{Radially Unbounded:} $\lim_{\|\boldsymbol{e}\| \to \infty} V(\boldsymbol{e}) = \infty$.
  \end{enumerate}
  
  These conditions ensure $V$ can be used to design stabilizing feedback controls, rendering the closed-loop system asymptotically stable at the origin. The CLF concept is not limited to quadratic forms but applies to any function meeting these criteria.
  
\end{theorem}

\begin{definition}[Safe State Space]
  Safe state space $\mathcal{B}$ is defined as:
  \begin{equation}
    \begin{aligned}
      \mathcal{B} = \big\{ \boldsymbol{e} \in \mathbb{R}^n \mid \exists \boldsymbol{u} \in \mathbb{U} \text{ such that } \\ \nabla V(\boldsymbol{e}) \cdot \left[f(\boldsymbol{x}^*) + g(\boldsymbol{x}^*)\boldsymbol{u}\right] \leq -\eta V(\boldsymbol{e}) \big\}
    \end{aligned}
  \end{equation}
  The safe state space $\mathcal{B}$ includes all states for which there exists at least one control input that ensures the CLF decreases at a sufficient rate to maintain safety and stability.
\end{definition}

\begin{definition}[Safe Energy Ball]
  Safe energy ball $\mathcal{D}$ is defined as the set
  \begin{equation}
    \mathcal{D} = \left\{ \boldsymbol{e} \mid V(\boldsymbol{e}) < V_0 \right\}  \text{ s.t. } \mathcal{D} \subseteq \mathcal{B}
  \end{equation}
  where $ V: \mathbb{R}^n \to \mathbb{R} $ is a control Lyapunov function (CLF), $\eta \in \mathbb{R}_{\geq 0}$ represents the coefficient of constraint, $V_0 \in \mathbb{R}_{>0}$ is associated with the size of the energy ball, and $\mathbb{U}$ denotes the set of admissible control inputs.
\end{definition}

\begin{figure}[htbp]
  \centering
  \includegraphics[width=0.35\textwidth]{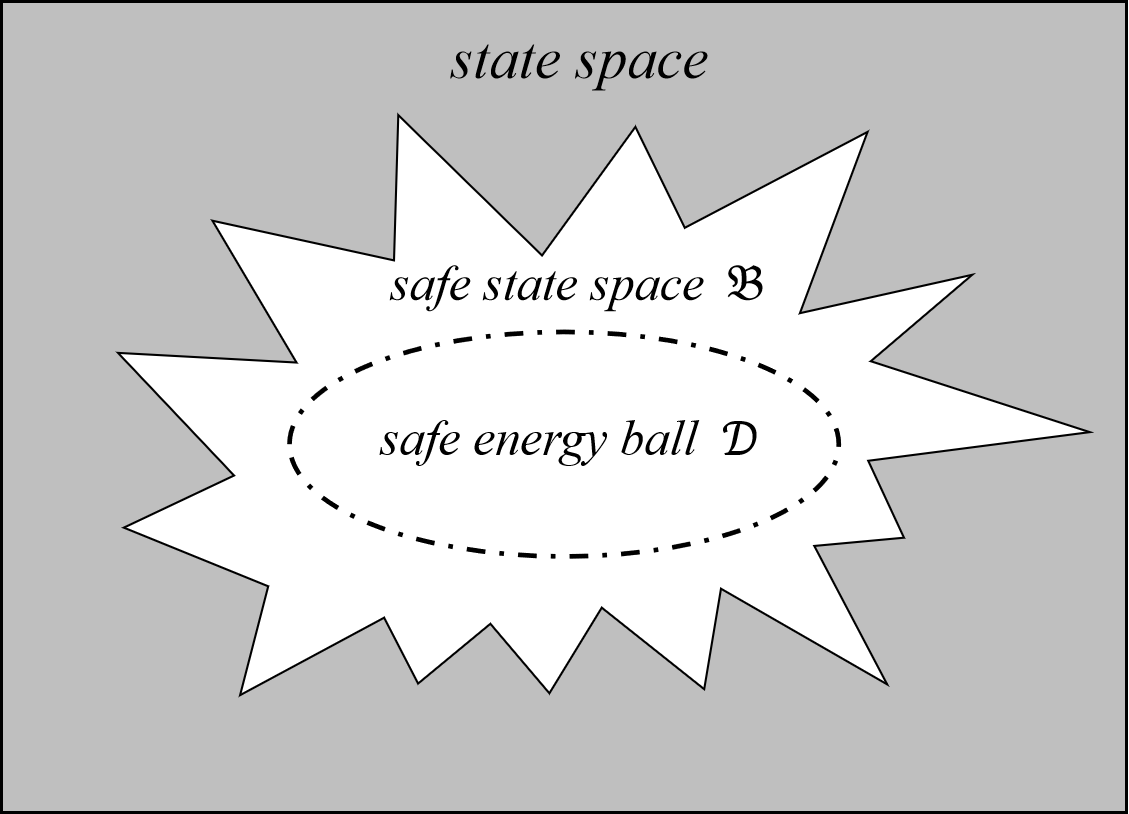}
  \caption{The hierarchical relationships: the state space contains all possible states; within it, the safe state space $\mathcal{B}$ includes states that ensure safety; and further within $\mathcal{B}$, the safe energy ball $\mathcal{D}$ comprises states with a control Lyapunov function below a threshold $V_0$.}
  \label{fig:safe_energy_ball}
\end{figure}

\begin{theorem}
  If the initial error $\boldsymbol{e}(0)$ is within the safe energy ball $\mathcal{D}$, then all subsequent errors $\boldsymbol{e}(t)$ for $t > 0$ will also remain within the safe energy ball $\mathcal{D}$. The proof can be shown as follows:
  \begin{enumerate}
    \item \textbf{Initial Condition:} Assume $\boldsymbol{e}(0) \in \mathcal{D}$. By definition, this means $\boldsymbol{e}(0)^T P \boldsymbol{e}(0) < V_0$ and there exists a control input $u$ such that $\nabla V(\boldsymbol{e}(0)) \cdot \left[f(\boldsymbol{x}^*(0)) + g(\boldsymbol{x}^*(0))u\right] \leq -\eta V(\boldsymbol{e}(0))$.

    \item \textbf{System Dynamics:} Consider the system dynamics described by the differential equation $\dot{\boldsymbol{e}}(t) = f(\boldsymbol{x}^*(t)) + g(\boldsymbol{x}^*(t))u$, where $f$ and $g$ are the uncontrolled and controlled dynamics, respectively, and $u$ is the control input.

    \item \textbf{Control Lyapunov Function (CLF):} Let $V(\boldsymbol{e})$ be a CLF that is positive definite and there exists a control law $u = k(\boldsymbol{e})$ such that $\dot{V}(\boldsymbol{e}) = \nabla V(\boldsymbol{e}) \cdot \left[f(\boldsymbol{x}^*) + g(\boldsymbol{x}^*)k(\boldsymbol{e})\right] \leq 0$.

    \item \textbf{Preservation of the Safe Energy Ball:} Since $V(\boldsymbol{e}(0)) < V_0$ and $\dot{V}(\boldsymbol{e}) \leq 0$, it follows that $V(\boldsymbol{e}(t)) \leq V(\boldsymbol{e}(0)) < V_0$ for all $t \geq 0$. If the CLF decrease condition is satisfied at $\boldsymbol{e}(0)$, and the system dynamics preserve this property, then it is satisfied for all $\boldsymbol{e}(t)$.
  \end{enumerate}

  Therefore, $\boldsymbol{e}(t) \in \mathcal{D}$ for all $t \geq 0$, ensuring that if the initial error is within the safe energy ball, all subsequent errors will also remain within the safe energy ball.
\end{theorem}

To design a Control Lyapunov Function (CLF) for nominal dynamic model, we start with the local linear model obtained by linearizing the system around an equilibrium point:
\begin{equation}
\dot{\boldsymbol{x}}^*(t) = \boldsymbol{A} \boldsymbol{x}^*(t) + \boldsymbol{B} \boldsymbol{u}(t)
\end{equation}
where $\boldsymbol{A} = \frac{\partial \boldsymbol{f}}{\partial \boldsymbol{x}^*} \bigg|_{\boldsymbol{x}^*=\boldsymbol{x}^*_d}$ and $\boldsymbol{B} = \boldsymbol{g}(\boldsymbol{x}^*_d)$ represent the local linear dynamic of nominal dynamic model evaluated at the desired state $\boldsymbol{x}^*_d$. This nominal dynamic model serves as the basis for applying Linear Quadratic Regulator (LQR) techniques to derive a positive definite matrix $\boldsymbol{P}$, used in constructing the quadratic CLF $V(\boldsymbol{e}) = \boldsymbol{e}^\top \boldsymbol{P} \boldsymbol{e}$. The nominal model simplifies CLF application and ensures stability across operating conditions.

To simplify the cost function for optimization, we apply a second-order Taylor expansion around the equilibrium point, neglecting first-order and higher-order (third-order and above) terms. This yields a purely quadratic approximation of the performance index:
\begin{equation}
J(\boldsymbol{e}, \boldsymbol{u}) = \int_0^\infty \left[ \boldsymbol{e}^{*}(t)^\top \boldsymbol{Q} \boldsymbol{e}(t) + \boldsymbol{u}^\top(t) \boldsymbol{R} \boldsymbol{u}(t) \right] dt
\end{equation}
where $\boldsymbol{Q} = \frac{\partial^2 J}{\partial z_i \partial z_j}\big|_{z_d}$ and $\boldsymbol{R} = \frac{\partial^2 J}{\partial u_i \partial u_j}$ are positive definite matrices that weigh the error $\boldsymbol{e}$ and control effort $\boldsymbol{u}$, respectively. To design an LQR controller, we aim to minimize the following quadratic cost function:
\begin{equation}
J = \int_{0}^{\infty} [\boldsymbol{e}(t)^\top \boldsymbol{Q} \boldsymbol{e}(t) + \boldsymbol{u}^T(t) \boldsymbol{R} \boldsymbol{u}(t)] dt
\end{equation}
where $\boldsymbol{Q}$ is a positive semi-definite state weighting matrix, and $\boldsymbol{R}$ is a positive definite control weighting matrix.

The optimal control policy near the equilibrium point is obtained by solving the Algebraic Riccati Equation (ARE):
\begin{equation}
\boldsymbol{A}^T \boldsymbol{P} + \boldsymbol{P}\boldsymbol{A} - \boldsymbol{P}\boldsymbol{B}\boldsymbol{R}^{-1}\boldsymbol{B}^T \boldsymbol{P} + \boldsymbol{Q} = 0
\label{eq:ARE}
\end{equation}
where $\boldsymbol{P}$ is the symmetric positive definite solution matrix. This setup ensures that $V(\boldsymbol{e}) = \boldsymbol{e}^\top \boldsymbol{P} \boldsymbol{e}$ and its time derivative $\dot{V}(\boldsymbol{e}) = -\boldsymbol{e}^\top \boldsymbol{Q} \boldsymbol{e} \leq 0$ satisfy the Lyapunov stability conditions, guaranteeing local asymptotic stability around the equilibrium point and optimality with respect to the given cost function.

While the LQR-based control policy is designed to be optimal in the vicinity of the equilibrium point, the associated Lyapunov function $ V(\boldsymbol{e}) = \boldsymbol{e}^\top \boldsymbol{P} \boldsymbol{e} $ can also be used to evaluate and ensure stability over a wider region of the state space.

\begin{figure}[htbp]
  \centering
  \includegraphics[width=0.35\textwidth]{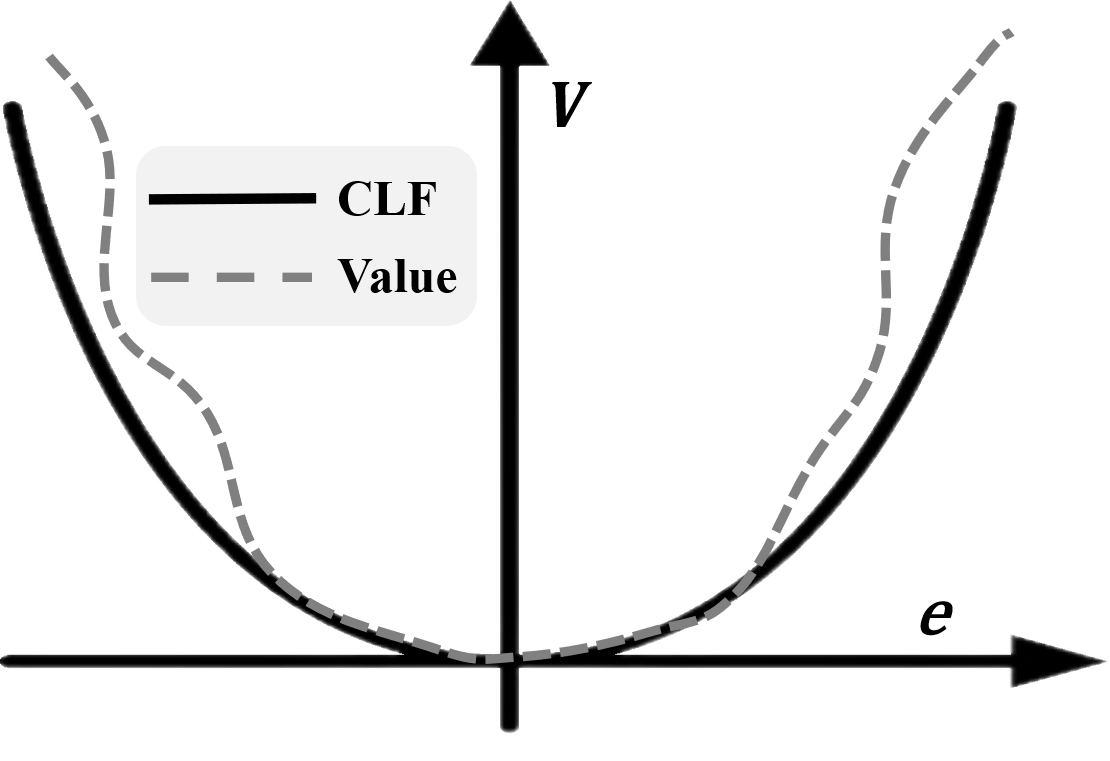}
  \caption{The CLF from an LQR-based control policy ensures both stability and local optimality, with the CLF and value function closely matching near the equilibrium.}
  \label{fig:LQR_CLF}
\end{figure}

\begin{figure}[htbp]
  \centering
  \includegraphics[width=0.35\textwidth]{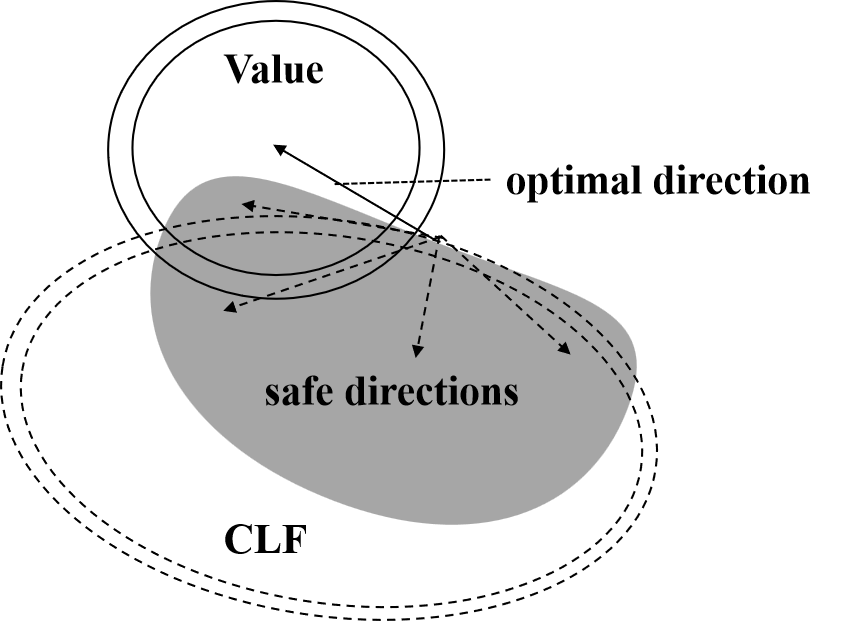}
  \caption{Optimal control input may not satisfy CLF constraints, compromising optimality.}
  \label{fig:safe_vs_opt}
\end{figure}

Optimal control input may not satisfy CLF constraints, compromising optimality. Figure.~\ref{fig:safe_vs_opt} illustrates how the optimal direction (based on value) may not always lie within the safe direction set defined by the Control Lyapunov Function (CLF), potentially leading to compromised optimality and stability.

\subsection{Adaptive Constraint Strength}

To tackle the issues of system stability and performance optimization in the face of unmodeled dynamics, a method for real-time adjustment of constraint conservatism has been devised. In control systems, uncertainties can significantly impact the dynamic behavior and robustness, thus this approach seeks to adaptively manage the level of constraint conservatism to maintain stability and performance under such conditions. The method involves ongoing assessment of the discrepancy $\delta(t)$ between the actual rate of change of the Lyapunov function $\dot{V}_{\text{actual}}(t)$ and its prescribed counterpart $\dot{V}_{\text{desired}}(t)$, ensuring a stable system response and optimal performance despite encountering unknown dynamics.

\begin{equation}
  \delta(t) = \dot{V}_{\text{desired}} - \dot{V}_{\text{actual}}
  \label{eq:error}
\end{equation}
serves as an indicator of the system's susceptibility to these dynamics. By assessing $\delta(t)$, the system can adaptively adjust the conservatism of its constraints: enforcing stricter limitations during significant disturbances to safeguard stability, while relaxing them in more stable states to enhance performance optimization. This adaptive modulation ensures that the system remains robust and efficient even when faced with dynamic uncertainties.

The adjustment factor $\eta(t)$ is defined through a time-varying multiplicative enhancement of the baseline conservatism $\eta_0$, given by:
\begin{equation}
  \eta(t) = \eta_0 \left(1 + k_{\eta}(t)\right)
\end{equation}
where $\eta_0$ represents the nominal level of constraint conservatism, and $k_{\eta}(t)$ denotes a gain coefficient that undergoes dynamic adaptation.

To ensure the system's stability and performance, the actual and desired constraints are formulated based on the Lyapunov function $V(\boldsymbol{x}(t))$, which characterizes the energy landscape of the system. The actual constraint, which accounts for dynamic uncertainties $\boldsymbol{d}(\boldsymbol{x}(t))$ and is given by Equation \ref{eq:actual_constraint}, is:
\begin{equation}
  \begin{aligned}
    \frac{\partial V}{\partial \boldsymbol{e}}& \left[\boldsymbol{f}(\boldsymbol{x}(t)) + \boldsymbol{g}(\boldsymbol{x}(t))\boldsymbol{u}(t) + \boldsymbol{d}(\boldsymbol{x}(t)) - \dot{\boldsymbol{x}}_d\right] \\< &- \eta_0 \left[1 + k_{\eta}(t)\right] V(\boldsymbol{e}(t))
  \end{aligned}
  \label{eq:actual_constraint}
\end{equation}
The desired constraint, assuming nominal system and presented in Equation \ref{eq:desired_constraint}, is:
\begin{equation}
  \frac{\partial V}{\partial \boldsymbol{e}} \left[\boldsymbol{f}(\boldsymbol{x}(t)) + \boldsymbol{g}(\boldsymbol{x}(t))\boldsymbol{u}(t) - \dot{\boldsymbol{x}}_d \right] < - \eta_0 V(\boldsymbol{e}(t))
  \label{eq:desired_constraint}
\end{equation}
The error term $\delta(t)$ quantifies the impact of model uncertainties on the system dynamics and is defined as:
\begin{equation}
  \delta(t) = -\frac{\partial V}{\partial \boldsymbol{e}} \boldsymbol{d}(\boldsymbol{x}(t))
\end{equation}
where $\boldsymbol{d}(\boldsymbol{x}(t))$ represents the model uncertainties. To align the actual constraints with the desired constraints, it is necessary to establish a relationship between the error term $e(t)$ and the system's state, specifically:
\begin{equation}
  -\frac{\partial V}{\partial \boldsymbol{e}} \boldsymbol{d}(\boldsymbol{x}(t)) = \eta_0 k_{\eta}(t) V(\boldsymbol{x}(t))
\end{equation}
This condition implies that the error term $e(t)$ must be proportional to the Lyapunov function $V(\boldsymbol{x}(t))$ and the adaptive parameter $k_{\eta}(t)$. By appropriately adjusting $k_{\eta}(t)$, the additional term $\eta_0 k_{\eta}(t) V(\boldsymbol{x}(t))$ can compensate for the model uncertainties, ensuring that the actual constraints match the desired constraints. Formally, if a suitable $k_{\eta}(t)$ can be found such that this condition holds, then the satisfaction of the desired constraints:
\begin{equation}
  \frac{\partial V}{\partial \boldsymbol{e}} \left[\boldsymbol{f}(\boldsymbol{x}(t)) + \boldsymbol{g}(\boldsymbol{x}(t))\boldsymbol{u}(t) - \dot{\boldsymbol{x}}_d\right] < - \eta_0 V(\boldsymbol{e}(t))
\end{equation}
automatically ensures the satisfaction of the actual constraints:
\begin{equation}
  \begin{aligned}
    \frac{\partial V}{\partial \boldsymbol{e}}& \left[\boldsymbol{f}(\boldsymbol{x}(t)) + \boldsymbol{g}(\boldsymbol{x}(t))\boldsymbol{u}(t) + \boldsymbol{d}(\boldsymbol{x}(t))  - \dot{\boldsymbol{x}}_d\right] \\< &- \eta_0 (1 + k_{\eta}(t)) V(\boldsymbol{e}(t))
  \end{aligned}
\end{equation}

\begin{equation}
\dot{k}_{\eta}(t) = \left(\frac{\delta(t)}{\eta_0 V(\boldsymbol{e}(t)) + \epsilon} - k_{\eta}(t)\right) \omega_{\eta}
\end{equation}
where $\epsilon > 0$ prevents division by zero (e.g., $\epsilon = 0.001$), and $\omega_{\eta} > 0$ modulates the adaptation speed (e.g., $\omega_{\eta} = 0.01$).The transfer function $H(s)$ derived from the Laplace transform is:
\begin{equation}
  H(s) = \frac{K_{\eta}(s)}{E(s)} = \frac{\omega_{\eta}}{(s + \omega_{\eta})(\eta_0 V(\boldsymbol{E}(s)) + \epsilon)}
\end{equation}
The poles of $H(s)$ are:
\begin{align}
  s_1 &= - \omega_{\eta} \\
  s_2 &= -\frac{\epsilon}{\eta_0 V(\boldsymbol{E}(s))}
\end{align}
Both poles lie in the left half-plane, ensuring the system's stability. The gain at low frequencies is given by:
\begin{equation}
  H(0) = \frac{\omega_{\eta}}{\omega_{\eta} (\eta_0 V(\boldsymbol{E}(s)) + \epsilon)} = \frac{1}{(\eta_0 V(\boldsymbol{E}(s)) + \epsilon)}
\end{equation}

At high frequencies, the gain approaches zero:
\begin{equation}
  H(\infty) = 0
\end{equation}

This mechanism ensures the system responds to low-frequency errors while effectively suppressing high-frequency noise.  The constraint conservatism factor $\eta(t) = \eta_0 (1 + k_{\eta}(t))$ is adjusted based on the error term $\delta(t)$.  When $\delta(t)$ is significant, $k_{\eta}(t)$ increases to reduce the discrepancy; when $\delta(t)$ is small, the adjustment rate slows to maintain stability. This adaptive approach balances robustness and performance, ensuring resilience to environmental changes and maintaining system stability under varying conditions.

\subsection{Safety-Prioritized Control Input Smoothing}

The introduction of a vibration-dampening term to the objective function significantly enhances control input smoothness while ensuring compliance with operational constraints. This approach distinguishes itself from both standard RL methods and simple inertia-based techniques by penalizing rapid changes in control actions and deviations from the RL-recommended inputs $\boldsymbol{u}^{RL}(t)$, thus mitigating vibrations and reducing unnecessary movements that can lead to actuator wear.

The modified objective function incorporating the vibration dampening effect is given by:
\begin{equation}
  \begin{aligned}
    J(\boldsymbol{u}) &= \int_0^T \bigg[ (\boldsymbol{u}(t) - \boldsymbol{u}^{RL}(t))^\top (\boldsymbol{u}(t) - \boldsymbol{u}^{RL}(t)) +\\& \underbrace{\beta (\boldsymbol{u}(t) - \boldsymbol{u}(t-\Delta t))^\top (\boldsymbol{u}(t) - \boldsymbol{u}(t-\Delta t))}_{\textbf{vibration-Dampening Term}}\bigg] dt
  \end{aligned}
\end{equation}
where the parameter $\beta$ controls the trade-off between closely following RL recommendations and ensuring smooth transitions. Unlike systems without the dampening term, which may exhibit abrupt changes in control actions, this formulation promotes gradual adjustments, leading to enhanced performance and durability.

When smooth actions are not required to meet constraints, the vibration-damper term's effect can be equivalently represented by:
\begin{equation}
\boldsymbol{u}(t) = \frac{\beta}{\beta + 1}\boldsymbol{u}(t-\Delta t) + \frac{1}{\beta + 1}\boldsymbol{u}^{RL}(t)
\end{equation}
where $\beta$ ensures that each control input $\boldsymbol{u}(t)$ is a weighted average of the previous input $\boldsymbol{u}(t-\Delta t)$ and the RL-recommended input $\boldsymbol{u}^{RL}(t)$. This formulation is equivalent to integrating a damping mechanism, maintaining smooth transitions while adhering to operational constraints. Higher $\beta$ values prioritize smoother transitions over rapid responses, ensuring stable and controlled operations.

\begin{figure}[htbp]
  \centering
  \includegraphics[width=0.45\textwidth]{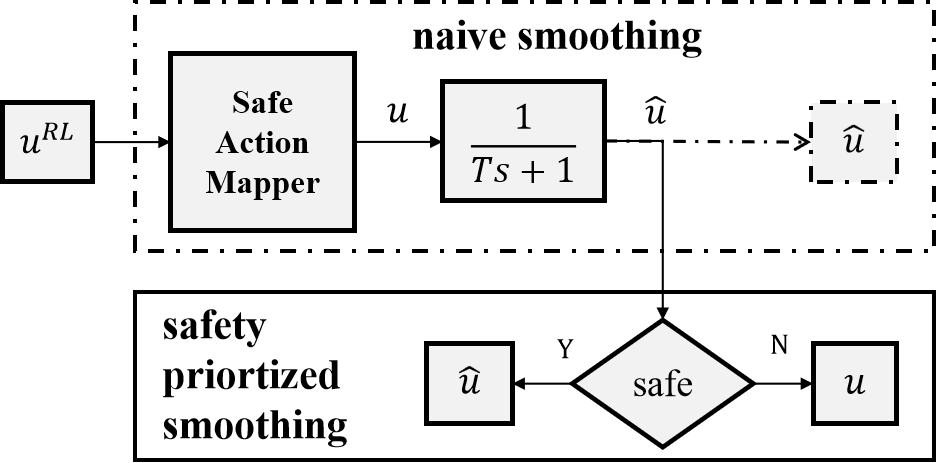}
  \caption{Safety-Prioritized Control Input Smoothing: Comparative Analysis Between the Proposed Method and Naive Smoothing Techniques}
  \label{fig:smooth_compare}
\end{figure}

Figure~\ref{fig:smooth_compare} provides a comparative analysis highlighting how the proposed method achieves smoother control transitions compared to naive smoothing techniques, while maintaining safety and stability. By integrating the vibration-dampening term, the control strategy not only aligns with RL recommendations but also ensures operational stability and safety, offering a robust solution for practical applications.

\subsection{Summary}
In this section, we present the formulation of a Quadratic Programming (QP) problem that integrates Control Lyapunov Function (CLF) constraints into the control strategy, alongside the pseudo-code for the Soft Actor-Critic with CLF (SAC-CLF) training and control algorithm. The goal is to ensure that the control actions not only follow the recommendations of the reinforcement learning (RL) policy but also maintain system stability and safety by adhering to actuator limits.
\begin{equation}
  \label{eq:QP_final}
  \begin{aligned}
  (\boldsymbol{u}(t), \varepsilon) =& \mathop{\arg\min}\limits_{\boldsymbol{u} \in U, \varepsilon \geq 0} \|\boldsymbol{u}(t) - \boldsymbol{u}^{\text{RL}}(t)\|_2^2 + \\ &\beta \|\boldsymbol{u}(t) - \boldsymbol{u}(t-\Delta t)\|_2^2 +  K_\varepsilon \varepsilon \\
  \text{s.t.} \quad &\frac{\partial V}{\partial \boldsymbol{e}} [\boldsymbol{f}(\boldsymbol{x}(t)) + \boldsymbol{g}(\boldsymbol{x}(t))\boldsymbol{u}(t)] \\ & \leq -\eta(t) V(\boldsymbol{e}(t)), \\
  & \boldsymbol{u}_{\text{low}} \leq \boldsymbol{u}(t) \leq \boldsymbol{u}_{\text{high}}
  \end{aligned}
\end{equation}

Where:
\begin{itemize}
  \item $ \boldsymbol{u}^{\text{RL}}(t) $ and $ \boldsymbol{u}(t-\Delta t) $: The control inputs generated by the reinforcement learning (RL) algorithm and from the previous time step.
  \item $ \eta(t) $: An adaptive parameter that adjusts the convergence rate. It increases if the Lyapunov function derivative $ \dot{V}(\boldsymbol{e}(t)) $ indicates instability, tightening the stability constraint, and decreases if the system performs well, relaxing the constraint. This ensures robustness and optimal performance.
  \item $ \boldsymbol{u}_{\text{low}} $ and $ \boldsymbol{u}_{\text{high}} $: The lower and upper bounds on the actuator limits.
  \item $ \varepsilon $ and $ K_\varepsilon $: A slack variable and a large positive penalty coefficient that allow a small, controlled violation of the CLF condition while heavily penalizing any such violations.
  \item $ \beta $: A user-defined positive number that balances the trade-off between minimizing the difference with the RL control input and the previous control input.
  \item $ \boldsymbol{P} $: A positive definite matrix, typically obtained by solving the Algebraic Riccati Equation (ARE) associated with the Linear Quadratic Regulator (LQR). It is used to construct the Lyapunov function $ V(\boldsymbol{e}) = \boldsymbol{e}^\top \boldsymbol{P} \boldsymbol{e} $ to assess system stability.
\end{itemize}

By solving the QP problem presented in Equation (\ref{eq:QP_final}), the resulting control input $ \boldsymbol{u}(t) $ aims to closely match the RL-suggested action while preserving system stability and respecting actuator limitations. This approach facilitates a balanced control strategy that prioritizes safety, performance, and transition smoothness.

Algorithm \ref{alg:SAC-CLF-Control} outlines the process for training and applying the SAC-CLF method. It starts with initializing a replay buffer and setting up parameters specific to the SAC algorithm, such as learning rates and target entropy, followed by defining parameters required for the CLF optimization problem. Each episode begins with environment initialization and proceeds through discrete time steps where the following operations occur:
\begin{algorithm}
  \renewcommand{\algorithmicrequire}{\textbf{Input:}}
  \renewcommand{\algorithmicensure}{\textbf{Output:}}
  \caption{SAC-CLF Training and Control Algorithm}
  \label{alg:SAC-CLF-Control}
  \begin{algorithmic}[1]
      \REQUIRE
      \STATE \quad Replay buffer $ \mathcal{D} $,
      \STATE \quad Learning rates for policy, Q-function, and value function $ \alpha_{\pi}, \alpha_{Q}, \alpha_{V} $,
      \STATE \quad Target entropy for policy optimization $ -dim(\boldsymbol{a}) $.
      
      \STATE \quad Initial state $ \boldsymbol{x}(0) $,
      \STATE \quad Initial control input from the previous time step $ \boldsymbol{u}(-\Delta t) $,
      \STATE \quad RL policy $ \pi^{\text{RL}} $,
      \STATE \quad Positive definite matrix $ \boldsymbol{P} $,
      \STATE \quad Actuator limits $ \boldsymbol{u}_{\text{low}}, \boldsymbol{u}_{\text{high}} $,
      \STATE \quad Penalty coefficient for slack variable $ K_\varepsilon $,
      \STATE \quad Trade-off parameter between smoothness and adherence to RL policy $ \beta > 0 $,
      \STATE \quad Adaptive convergence rate parameter $ \eta(t) > 0 $.
      
      \ENSURE Optimal control law $\pi(\cdot)$.
      \STATE Initialize replay buffer $ \mathcal{D} $.
      \FOR{each episode}
          \STATE Initialize the environment and get the initial state $ \boldsymbol{x}(0) $.
          \FOR{each time step $ t = 0, \Delta t, 2\Delta t, \ldots $ within the episode}
              \STATE Obtain the RL control input $ \boldsymbol{u}^{\text{RL}}(t) = \pi^{\text{RL}}(\boldsymbol{x}(t)) $ with added Gaussian noise for exploration.
              \STATE Formulate and solve the QP as described in Equation.~(\ref{eq:QP_final})
              \STATE Apply the control input $ \boldsymbol{u}(t) $ to the system.
              \STATE Observe the next state $ \boldsymbol{x}(t+\Delta t) $ and reward $ r(t) $.
              \STATE Store the transition $(\boldsymbol{x}(t), \boldsymbol{u}(t), r(t), \boldsymbol{x}(t+\Delta t))$ into $ \mathcal{D} $.
              \STATE Sample a batch of transitions from $ \mathcal{D} $.
              \STATE Update the critic networks using Bellman error of the sampled transitions.
              \STATE Update the actor network by minimizing the expected entropy-regularized KL divergence.
              \STATE Update the temperature parameter to maintain the desired level of entropy.
              \STATE Update the adaptive parameter $ \eta(t) $ based on the system's performance.
              \STATE Perform soft updates for the target networks.
          \ENDFOR
      \ENDFOR
  \end{algorithmic}
\end{algorithm}

\section{Simulation}
In this section, we aim to evaluate the effectiveness of the proposed method through a series of simulation experiments.  Specifically, we perform three studies within the contexts of classical NCT system \cite{vamvoudakis_online_2010} and satellite attitude control: (1) We validate the efficacy of our Control Lyapunov Function (CLF) design by comparing it with a unit matrix CLF using Linear Quadratic Regulator (LQR) techniques.(2) We investigate the capability of adaptive constraints to strike a balance between system robustness and performance.(3) We examine the impact of the vibration-dampening term on control input smoothing.

  \subsection{Environments}

  \textbf{The NCT system}
    simulates a 2D dynamical system. The state of the system is a 2D vector $\boldsymbol{x} = [x_1, x_2]^T$, and the control input is a scalar $u$. The dynamics are defined by the functions:
    \begin{equation}
      \begin{aligned}
        \begin{bmatrix} \dot{x}_1 \\ \dot{x}_2 \end{bmatrix} =& 
      \underbrace{\begin{bmatrix} -x_1 + x_2 \\ -0.5 x_1 - 0.5 x_2 (1 - (\cos(2 x_1) + 2)^2) \end{bmatrix}}_{f(\boldsymbol{x})} +\\ &
      \underbrace{\begin{bmatrix} 0 \\ \cos(2 x_1) + 2 \end{bmatrix}}_{g(\boldsymbol{x})} u
      \end{aligned}
    \end{equation}
  The control objective is to stabilize the system at the origin $(0, 0)$ while minimizing the control effort and penalizing large state deviations. The environment provides a continuous action space $u \in [-10.0, 10.0]$ and a 2D continuous observation space for the state. The cost function for the NCT system is defined as:
  \begin{equation}
    J_{\text{NCT}} = \int_0^T (x_1^2 + x_2^2) \, dt
  \end{equation}
  
  \textbf{The satellite attitude control environment}
    simulates the rotational dynamics of a satellite using quaternions and angular velocities. The state is represented by a 7-dimensional vector $[q_0, q_1, q_2, q_3, \omega_x, \omega_y, \omega_z]^T$, where $q_0, q_1, q_2, q_3$ are the components of the quaternion representing the orientation error, and $\omega_x, \omega_y, \omega_z$ are the components of the angular velocity vector. The action space consists of three components representing the applied torques, each bounded within $[-0.5, 0.5]^3$. The dynamics are governed by the equations:
    \begin{equation}
      \dot{\boldsymbol{x}} = 
      \underbrace{\begin{bmatrix} 
          \frac{1}{2} \boldsymbol{R}(\boldsymbol{q}) \bm{\omega} \\
          \boldsymbol{\boldsymbol{I}^{-1} \bm{\omega}\boldsymbol{I} \bm{\omega}}
      \end{bmatrix}}_{f(\boldsymbol{x})} +
      \underbrace{\begin{bmatrix} 
          \boldsymbol{0} \\
          \boldsymbol{I}^{-1} 
      \end{bmatrix}}_{g(\boldsymbol{x})}\boldsymbol{u}
  \end{equation}
  The control objective is to align the spacecraft's attitude with a desired orientation by driving the orientation error quaternion to $[0, 0, 0, 1]$ and reducing the angular velocity to zero. The cost function for the satellite attitude control is defined as:
  \begin{equation}
    J_{\text{satellite}} = \int_0^T (q_0^2 + q_1^2 + q_2^2 + \omega_x^2 + \omega_y^2 + \omega_z^2) \, dt
  \end{equation}

  \subsection{Validation of task-specific CLF Design}
  To illustrate the validity of our CLF design, we compare the performance of our customized CLF against a unit matrix CLF using LQR techniques. This comparative analysis provides insights into the advantages and potential limitations of our tailored CLF compared to a conventional approach.

  \begin{table}[ht]
    \centering
    \caption{Hyperparameters for simulations validating our CLF design (without model bias, constraints adaptation, or command smoothing).}
    \begin{tabular}{lcccc}
        \toprule
        Environment & $\eta_0$ & $\omega_{\eta}$ & $K_{\epsilon}$ &$\beta$ \\
        \midrule
        NCT system & 0.1 & 0.0 & $10^8$ & 0.0\\
        Satellite & 0.1 & 0.0 & $10^8$ & 0.0\\
        \bottomrule
    \end{tabular}
    \label{tab:CLF_parameters}
  \end{table}
  \begin{figure}[ht]
    \centering
    \begin{subfigure}[t]{0.45\textwidth}
        \centering
        \includegraphics[width=\textwidth]{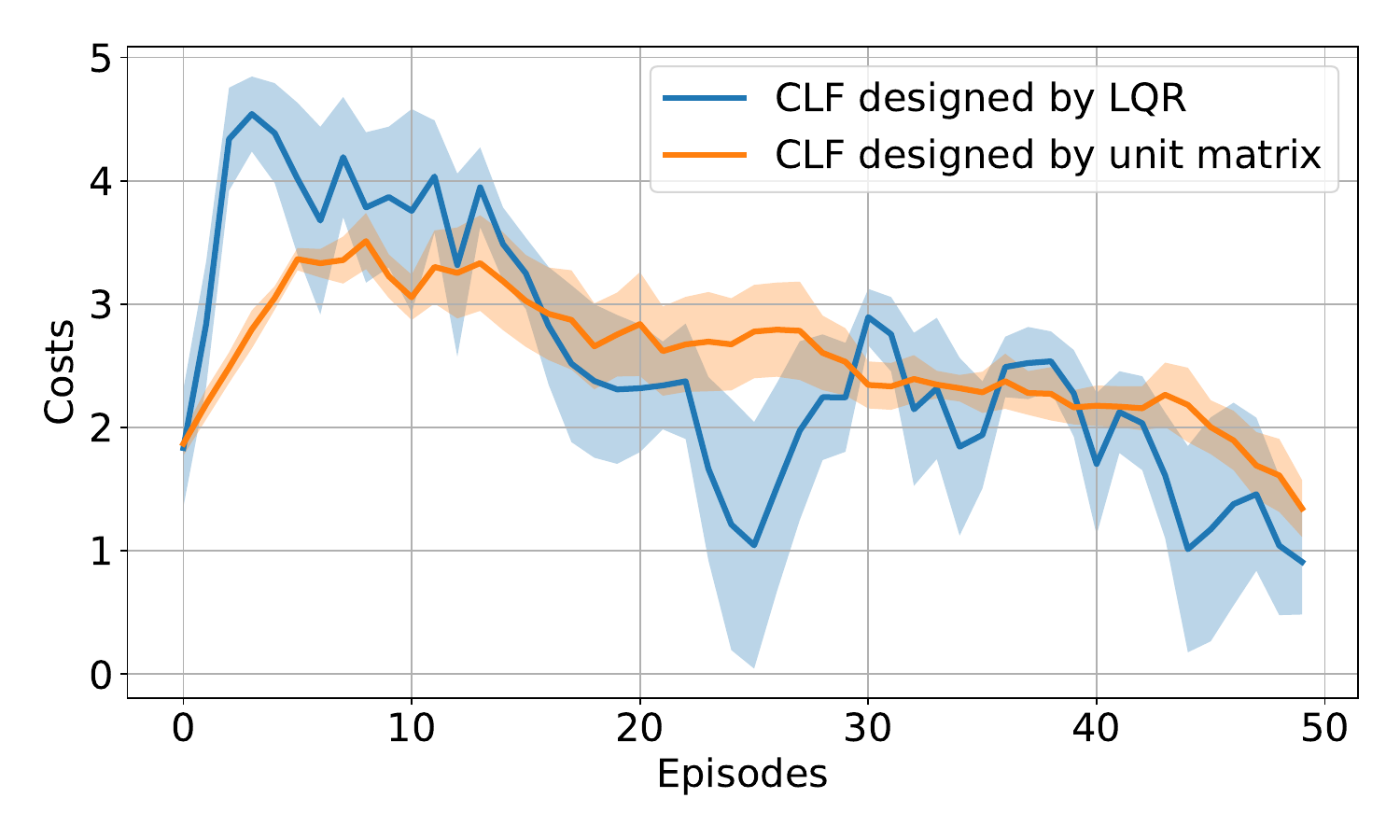}
        \caption{NCT System: \qquad $V_{LQR}(x_1, x_2) = 0.5 \cdot x_1^2 + x_2^2$ \qquad and $V_{unit}(x_1, x_2) = x_1^2 + x_2^2$}
    \end{subfigure}
    \hfill
    \begin{subfigure}[t]{0.45\textwidth}
        \centering
        \includegraphics[width=\textwidth]{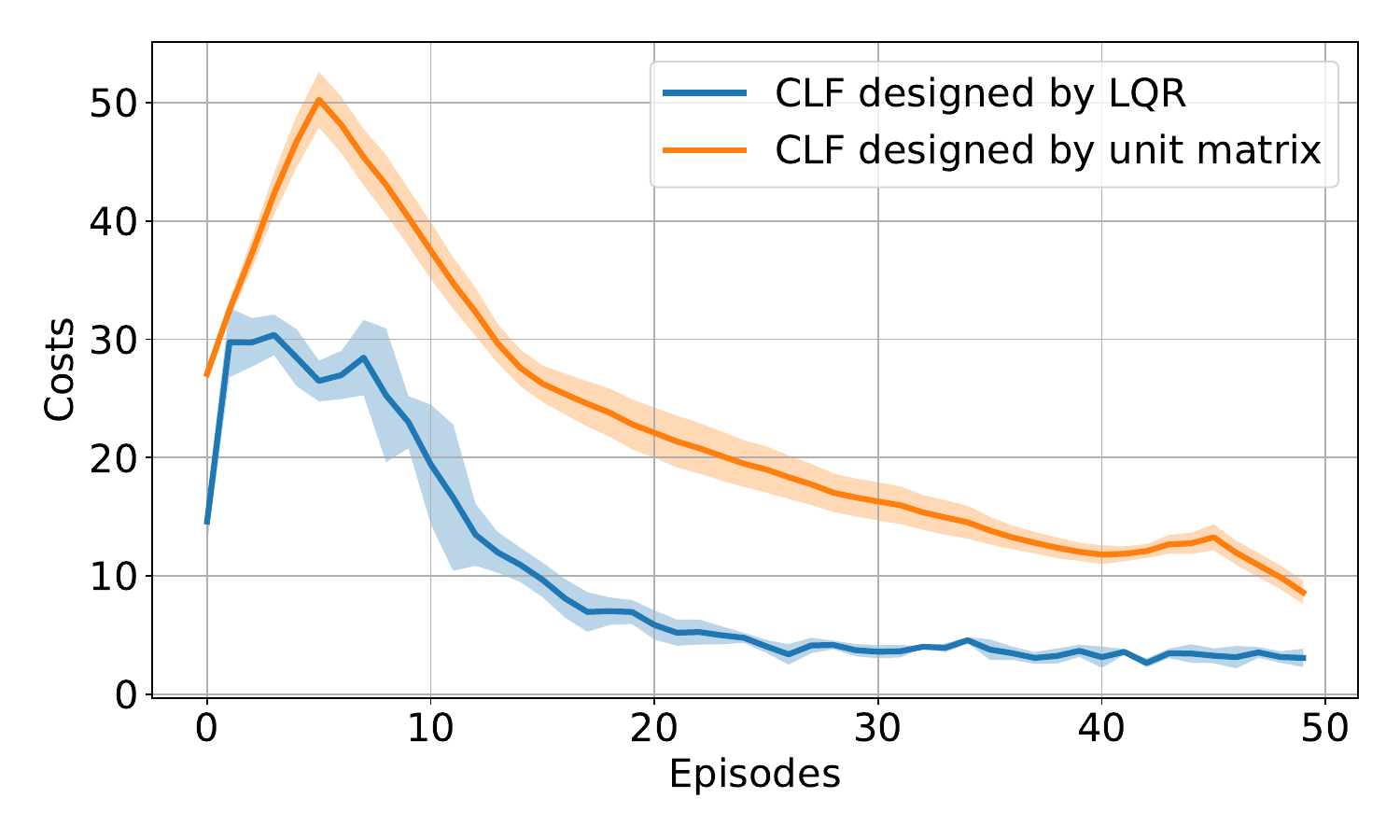}
        \caption{Satellite: $V_{LQR}(q_0, q_1, q_2, \omega_x, \omega_y, \omega_z) = 1.657 \cdot q_0^2 + 1.109 \cdot q_1^2 + 1.611 \cdot q_2^2 + 1.444 \cdot \omega_x^2 + 0.128 \cdot \omega_y^2 + 1.284 \cdot \omega_z^2 + 2 \cdot 0.872 \cdot q_0 \cdot \omega_x + 2 \cdot 0.115 \cdot q_1 \cdot \omega_y + 2 \cdot 0.797 \cdot q_2 \cdot \omega_z$ and $V_{unit}(q_0, q_1, q_2, \omega_x, \omega_y, \omega_z) = q_0^2 + q_1^2 + q_2^2 + \omega_x^2 + \omega_y^2 + \omega_z^2$}
    \end{subfigure}
    \caption{The customized CLF demonstrates superior convergence and stability compared to the unit matrix CLF. Each environment underwent 5 simulations, with solid lines representing the mean costs and shaded areas indicating the standard deviation.}
    \label{fig:CLFLQR}
  \end{figure}
  
  The results in Figure \ref{fig:CLFLQR} show that the customized CLF significantly impacts the cost metrics, but the effects differ between the NCT system and the satellite attitude control. In the NCT system, the costs associated with the two CLFs do not exhibit significant differences, suggesting that their performance is quite similar. However, for the satellite, the customized CLF leads to notably lower costs, indicating better performance in terms of precision and stability.

  \subsection{Validation of Adaptive Constraints Strength}
  To illustrate the ability of adaptive constraints to balance system robustness and performance, the final set of experiments compares scenarios with and without these constraints, highlighting how they enhance system stability while maintaining high performance standards.
  \begin{table}[ht]
    \centering
    \caption{Hyperparameters for simulations validating adaptive constraints in the presence of model bias (without command smoothing).}
    \begin{tabular}{lcccc}
        \toprule
        Environment & $\eta_0$ & $\omega_{\eta}$ & $K_{\epsilon}$ &$\beta$ \\
        \midrule
        \makecell{NCT system (strict\\and constant constraints)}  & 0.1 & 0.0 & $10^8$ & 0.0\\
        \makecell{NCT system (strict\\and adpative constraints)} & 0.1 & 0.01 & $10^8$ & 0.0\\
        \makecell{NCT system (slight\\and constant constraints)} & 0.01 & 0.0 & $10^8$ & 0.0\\
        \makecell{NCT system (slight\\and adpative constraints)} & 0.01 & 0.01 & $10^8$ & 0.0\\
        \makecell{Satellite (strict\\and adpative constraints)}  & 0.3 & 0.0 & $10^8$ & 0.0\\
        \makecell{Satellite (strict\\and constant constraints)} & 0.3 & 0.01 & $10^8$ & 0.0\\
        \makecell{Satellite (slight\\and adpative constraints)} & 0.03 & 0.0 & $10^8$ & 0.0\\
        \makecell{Satellite (slight\\and constant constraints)} & 0.03 & 0.01 & $10^8$ & 0.0\\
        \bottomrule
    \end{tabular}
    \label{tab:adaptation_parameters}
\end{table}

  \begin{figure}[ht]
    \centering
    \begin{subfigure}[b]{0.45\textwidth}
        \centering
        \includegraphics[width=\textwidth]{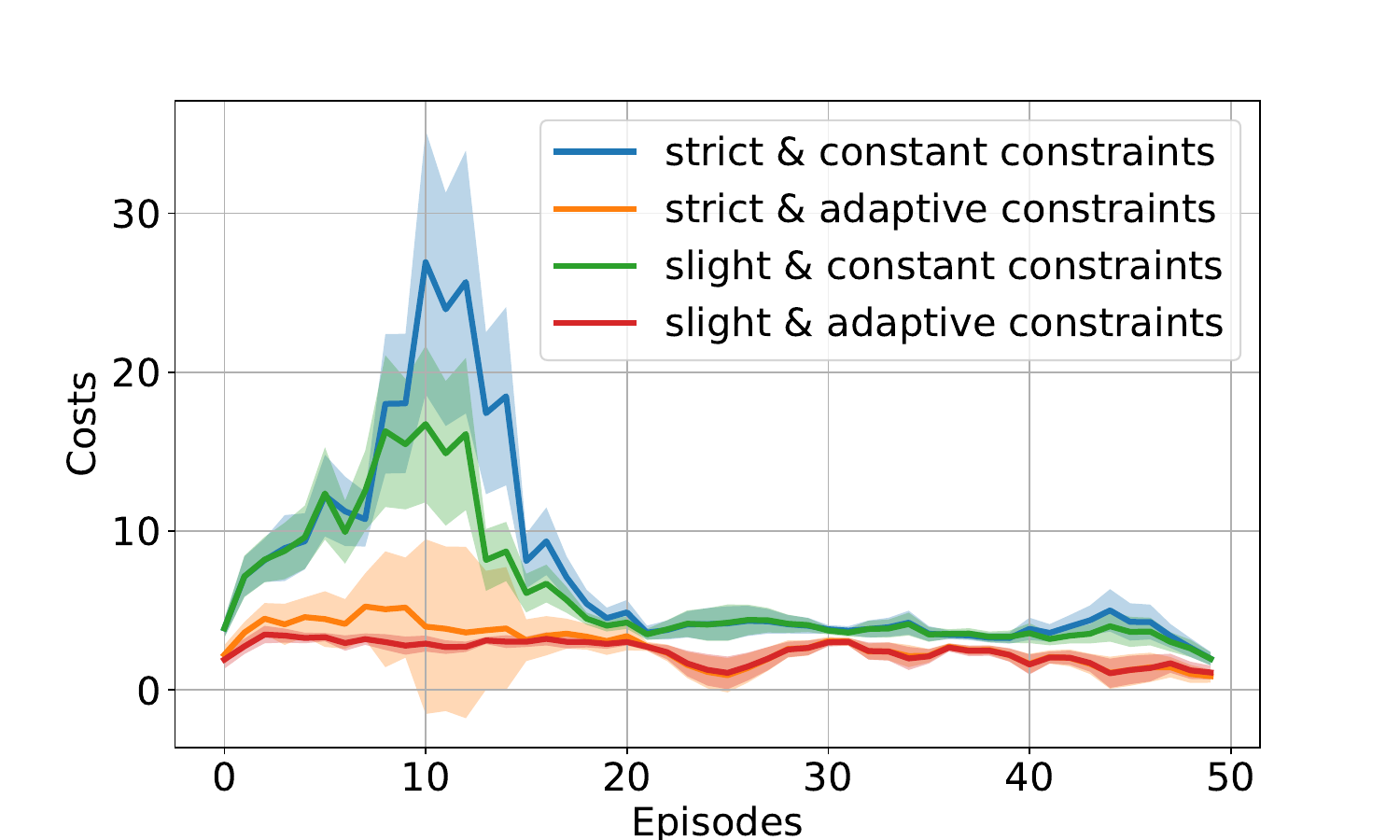}
        \caption{NCT System}
    \end{subfigure}
    \begin{subfigure}[b]{0.45\textwidth}
        \centering
        \includegraphics[width=\textwidth]{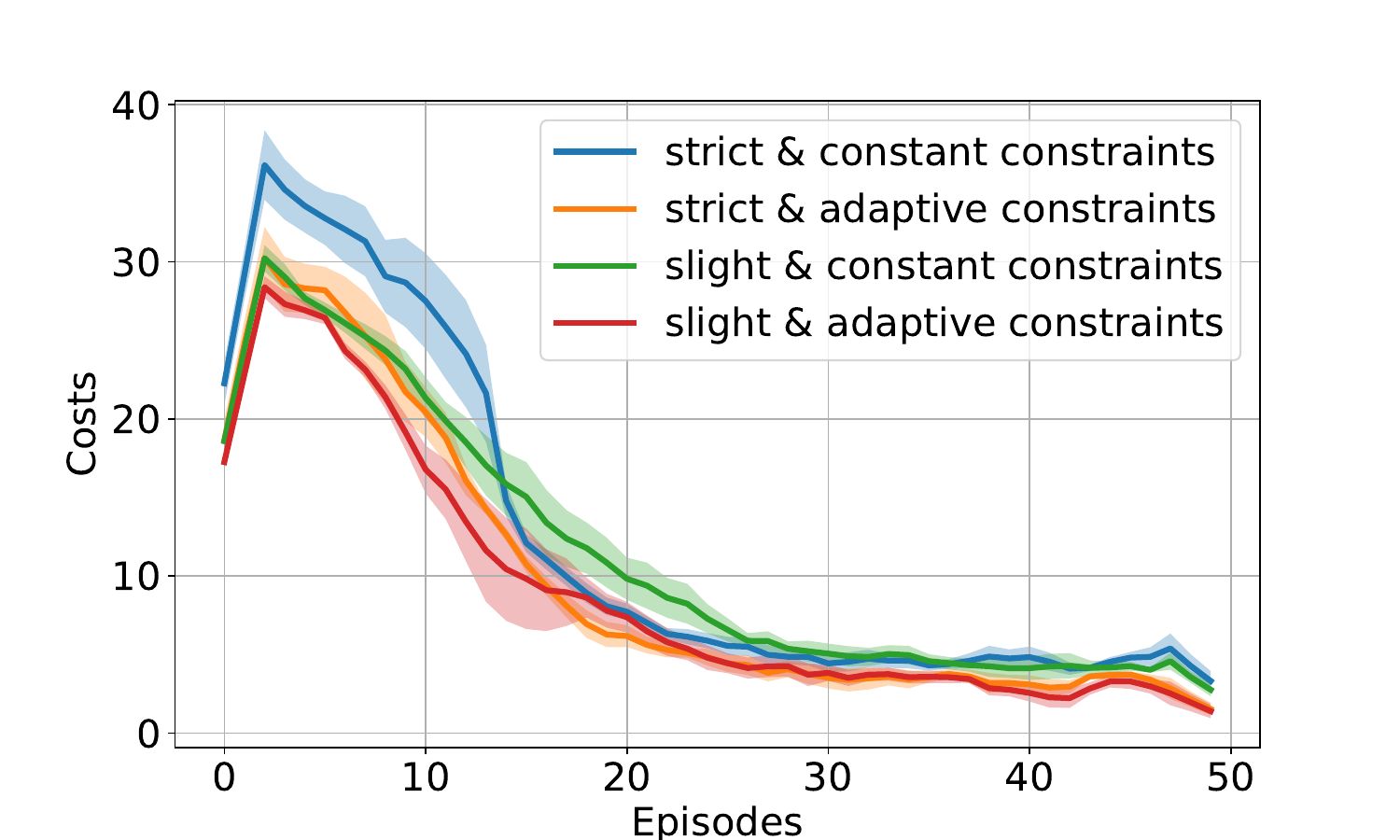}
        \caption{Satellite}
    \end{subfigure}
    \caption{Balancing robustness (constraints) and performance (costs) with adaptive constraints: (a) NCT System, (b) Satellite. Adaptive constraints improve system robustness and maintain high performance under varying conditions.}
    \label{fig:adaptation}
  \end{figure}
  The experimental results in Figure \ref{fig:adaptation} show that adaptive constraints have a more pronounced effect on reducing costs in the NCT system, where the mean costs are significantly lower and the standard deviation is smaller, indicating more consistent and stable performance. For the satellite attitude control, the reduction in mean costs is less significant, though there is still an improvement in performance stability.
  
  \subsection{Validation of Safety-Prioritized Control Input Smoothing}
  
  To illustrate the impact of the vibration-dampening term, we analyze the smoothness of commands both with and without this term, aiming to highlight its contribution to overall system stability and output quality.

  \begin{table}[ht]
    \centering
    \caption{Hyperparameters for simulations validating the vibration-dampening term in control input smoothing (without model bias or constraints adaptation).}
    \begin{tabular}{lcccc}
        \toprule
        Environment & $\eta_0$ & $\omega_{\eta}$ & $K_{\epsilon}$ &$\beta$ \\
        \midrule
        \makecell{NCT system (with \\vibration-dampening term)} & 0.1 & 0.0 & $10^8$ & 0.0\\
        \makecell{NCT system (without \\vibration-dampening term)} & 0.1 & 0.0 & $10^8$ & 1.0\\
        \makecell{Satellite (with \\vibration-dampening term)} & 0.1 & 0.0 & $10^8$ & 0.0\\
        \makecell{Satellite (without \\vibration-dampening term)}& 0.1 & 0.0 & $10^8$ & 1.0\\
        \bottomrule
    \end{tabular}
    \label{tab:smooth_parameters}
\end{table}
  
  \begin{figure}[ht]
    \centering
    \begin{subfigure}[t]{0.40\textwidth}
        \centering
        \includegraphics[width=\textwidth]{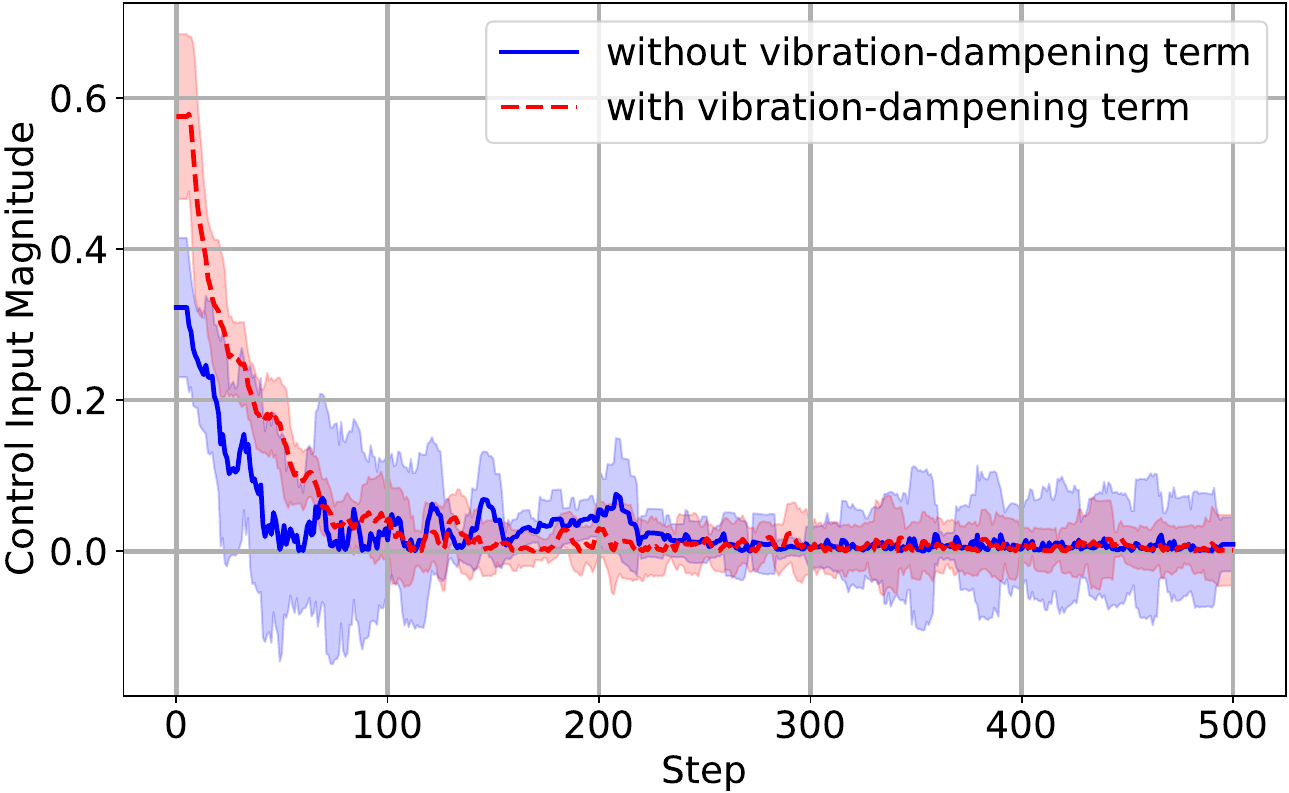}
        \caption{NCT System}
    \end{subfigure}
    \begin{subfigure}[t]{0.42\textwidth}
        \centering
        \includegraphics[width=\textwidth]{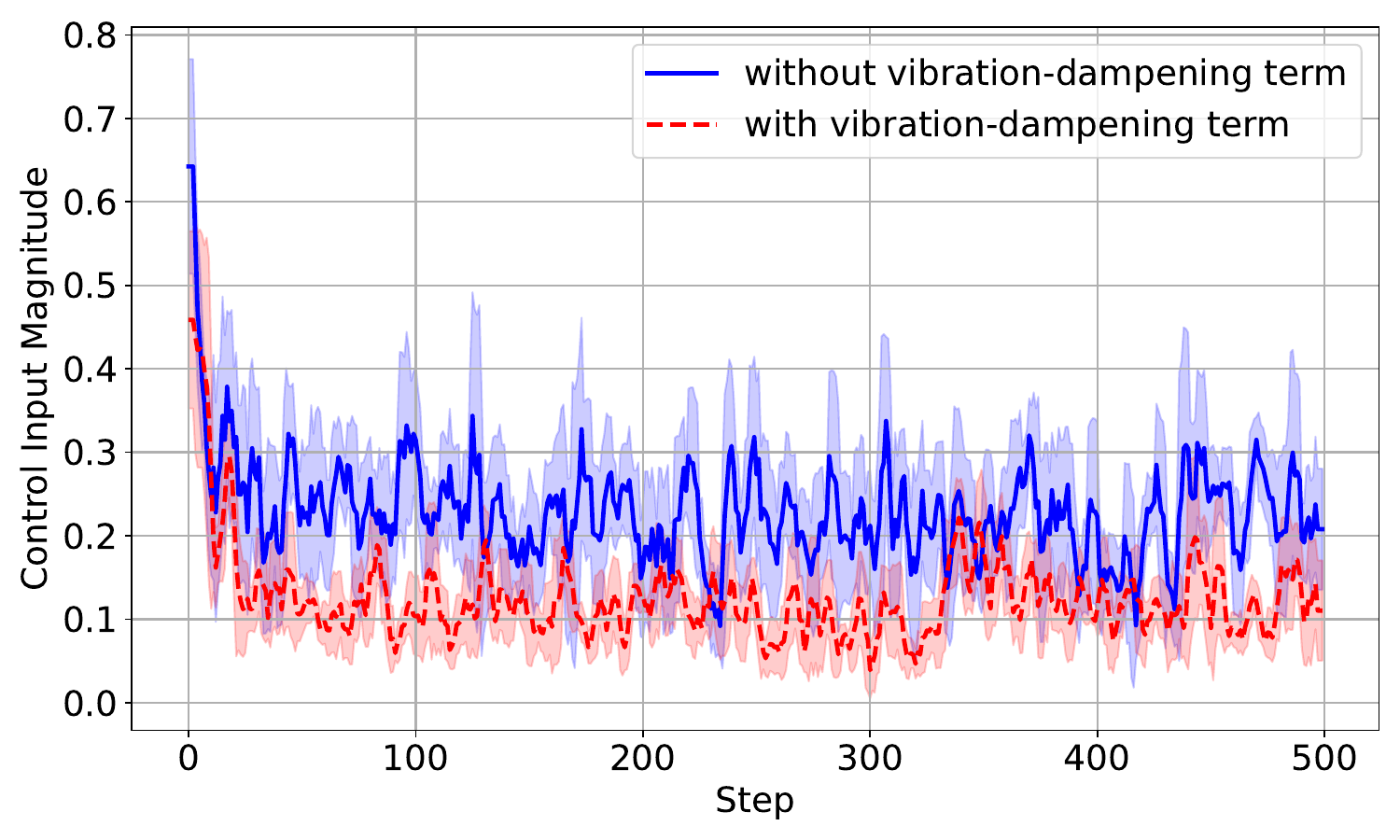}
        \caption{Satellite}
    \end{subfigure}
    \caption{Effectiveness of the vibration-dampening term in control input smoothing. The inclusion of the vibration-dampening term significantly reduces command oscillations and improves overall system stability.}
    \label{fig:smooth}
  \end{figure}
  
  Figure \ref{fig:smooth} demonstrates that the inclusion of the vibration-dampening term significantly reduces command oscillations and improves the smoothness of the control signals. In the NCT system, the dampened commands result in smoother state transitions and reduced wear on the system components. For the satellite, the dampened commands lead to more stable and precise attitude adjustments, minimizing jitter and improving the overall performance.

\section{Conclusion}
\label{sec:Conclusion}
SAC-CLF is an advanced controller architecture that merges model-free reinforcement learning (RL) with robust safety measures and dynamic adjustments. It introduces three key contributions to ensure safety during the RL process: (1) a systematic CLF design using linearization and LQR for stability, (2) a vibration dampening term to smooth control actions and reduce oscillations, and (3) adaptive constraint tuning to dynamically adjust constraints based on deviations from target CLF derivatives, enhancing both safety and performance.

Testing on nonlinear systems and satellite control has shown SAC-CLF to excel in sample efficiency and safety, outperforming other leading algorithms. This approach advances safe RL and paves the way for its application in physical system control, where reliability is critical. Future work will extend SAC-CLF to more complex scenarios and integrate it into safety-critical systems like autonomous vehicles and robotics.

\bibliographystyle{IEEEtran}
\bibliography{references}

\begin{thebibliography}{10}
\providecommand{\url}[1]{#1}
\csname url@samestyle\endcsname
\providecommand{\newblock}{\relax}
\providecommand{\bibinfo}[2]{#2}
\providecommand{\BIBentrySTDinterwordspacing}{\spaceskip=0pt\relax}
\providecommand{\BIBentryALTinterwordstretchfactor}{4}
\providecommand{\BIBentryALTinterwordspacing}{\spaceskip=\fontdimen2\font plus
\BIBentryALTinterwordstretchfactor\fontdimen3\font minus \fontdimen4\font\relax}
\providecommand{\BIBforeignlanguage}[2]{{%
\expandafter\ifx\csname l@#1\endcsname\relax
\typeout{** WARNING: IEEEtran.bst: No hyphenation pattern has been}%
\typeout{** loaded for the language `#1'. Using the pattern for}%
\typeout{** the default language instead.}%
\else
\language=\csname l@#1\endcsname
\fi
#2}}
\providecommand{\BIBdecl}{\relax}
\BIBdecl

\bibitem{zhao_-board_2023}
\BIBentryALTinterwordspacing
Y.~Zhao, H.~Yang, S.~Li, and Y.~Zhou, ``\BIBforeignlanguage{en}{On-board modeling of gravity fields of elongated asteroids using hopfield neural networks},'' \emph{\BIBforeignlanguage{en}{Astrodynamics}}, vol.~7, no.~1, pp. 101--114, Mar. 2023. [Online]. Available: \url{https://doi.org/10.1007/s42064-022-0151-3}
\BIBentrySTDinterwordspacing

\bibitem{li_closed-loop_2023}
\BIBentryALTinterwordspacing
W.~Li, Y.~Song, L.~Cheng, and S.~Gong, ``\BIBforeignlanguage{en}{Closed-loop deep neural network optimal control algorithm and error analysis for powered landing under uncertainties},'' \emph{\BIBforeignlanguage{en}{Astrodynamics}}, vol.~7, no.~2, pp. 211--228, Jun. 2023. [Online]. Available: \url{https://doi.org/10.1007/s42064-022-0153-1}
\BIBentrySTDinterwordspacing

\bibitem{wu_low-thrust_2024}
\BIBentryALTinterwordspacing
D.~Wu, L.~Cheng, S.~Gong, and H.~Baoyin, ``\BIBforeignlanguage{en}{Low-thrust trajectory optimization with averaged dynamics using analytical switching detection},'' \emph{\BIBforeignlanguage{en}{Journal Of Guidance, Control, And Dynamics}}, vol.~47, no.~6, pp. 1135--1149, Jun. 2024, publisher: American Institute of Aeronautics and Astronautics. [Online]. Available: \url{https://arc.aiaa.org/doi/10.2514/1.G007858}
\BIBentrySTDinterwordspacing

\bibitem{wang_spacecraft_2024}
\BIBentryALTinterwordspacing
W.~Wang, G.~Mengali, A.~A. Quarta, and H.~Baoyin, ``\BIBforeignlanguage{en}{Spacecraft relative motion control near an asteroid with uncertainties: a lyapunov redesign approach},'' \emph{\BIBforeignlanguage{en}{IEEE Transactions on Aerospace and Electronic Systems}}, vol.~60, no.~4, pp. 4507--4517, Aug. 2024, conference Name: IEEE Transactions on Aerospace and Electronic Systems. [Online]. Available: \url{https://ieeexplore.ieee.org/document/10476750}
\BIBentrySTDinterwordspacing

\bibitem{ionescu_robust_2020}
\BIBentryALTinterwordspacing
C.~M. Ionescu, E.~H. Dulf, M.~Ghita, and C.~I. Muresan, ``\BIBforeignlanguage{en}{Robust controller design: recent emerging concepts for control of mechatronic systems},'' \emph{\BIBforeignlanguage{en}{Journal of the Franklin Institute}}, vol. 357, no.~12, pp. 7818--7844, Aug. 2020. [Online]. Available: \url{https://www.sciencedirect.com/science/article/pii/S0016003220303999}
\BIBentrySTDinterwordspacing

\bibitem{qu_dynamic-matching_2024}
\BIBentryALTinterwordspacing
C.~Qu, L.~Cheng, S.~Gong, and X.~Huang, ``\BIBforeignlanguage{en}{Dynamic-matching adaptive sliding mode control for hypersonic vehicles},'' \emph{\BIBforeignlanguage{en}{Aerospace Science and Technology}}, vol. 149, p. 109159, Jun. 2024. [Online]. Available: \url{https://www.sciencedirect.com/science/article/pii/S127096382400292X}
\BIBentrySTDinterwordspacing

\bibitem{wang_adaptive_2017}
\BIBentryALTinterwordspacing
Y.~Wang and Q.~Wu, ``\BIBforeignlanguage{en}{Adaptive non-affine control for the short-period model of a generic hypersonic flight vehicle},'' \emph{\BIBforeignlanguage{en}{Aerospace Science and Technology}}, vol.~66, pp. 193--202, Jul. 2017. [Online]. Available: \url{https://www.sciencedirect.com/science/article/pii/S1270963816307933}
\BIBentrySTDinterwordspacing

\bibitem{lillicrap_continuous_2019}
\BIBentryALTinterwordspacing
T.~P. Lillicrap, J.~J. Hunt, A.~Pritzel, N.~Heess, T.~Erez, Y.~Tassa, D.~Silver, and D.~Wierstra, ``\BIBforeignlanguage{en}{Continuous control with deep reinforcement learning},'' Jul. 2019, arXiv:1509.02971. [Online]. Available: \url{http://arxiv.org/abs/1509.02971}
\BIBentrySTDinterwordspacing

\bibitem{schulman_trust_2017}
\BIBentryALTinterwordspacing
J.~Schulman, S.~Levine, P.~Moritz, M.~I. Jordan, and P.~Abbeel, ``Trust {Region} {Policy} {Optimization},'' Apr. 2017, arXiv:1502.05477 [cs]. [Online]. Available: \url{http://arxiv.org/abs/1502.05477}
\BIBentrySTDinterwordspacing

\bibitem{zhao_sim--real_2020}
\BIBentryALTinterwordspacing
W.~Zhao, J.~P. Queralta, and T.~Westerlund, ``Sim-to-{Real} {Transfer} in {Deep} {Reinforcement} {Learning} for {Robotics}: a {Survey},'' in \emph{2020 {IEEE} {Symposium} {Series} on {Computational} {Intelligence} ({SSCI})}, Dec. 2020, pp. 737--744, arXiv:2009.13303 [cs]. [Online]. Available: \url{http://arxiv.org/abs/2009.13303}
\BIBentrySTDinterwordspacing

\bibitem{ohnishi_safety-aware_2018}
\BIBentryALTinterwordspacing
M.~Ohnishi, L.~Wang, G.~Notomista, and M.~Egerstedt, ``\BIBforeignlanguage{en}{Safety-aware adaptive reinforcement learning with applications to brushbot navigation},'' Jan. 2018, arXiv:1801.09627 version: 1. [Online]. Available: \url{http://arxiv.org/abs/1801.09627}
\BIBentrySTDinterwordspacing

\bibitem{shah_airsim_2017}
\BIBentryALTinterwordspacing
S.~Shah, D.~Dey, C.~Lovett, and A.~Kapoor, ``{AirSim}: {High}-{Fidelity} {Visual} and {Physical} {Simulation} for {Autonomous} {Vehicles},'' Jul. 2017, arXiv:1705.05065 [cs]. [Online]. Available: \url{http://arxiv.org/abs/1705.05065}
\BIBentrySTDinterwordspacing

\bibitem{ames_control_2017}
\BIBentryALTinterwordspacing
A.~D. Ames, X.~Xu, J.~W. Grizzle, and P.~Tabuada, ``\BIBforeignlanguage{en}{Control {Barrier} {Function} {Based} {Quadratic} {Programs} for {Safety} {Critical} {Systems}},'' \emph{\BIBforeignlanguage{en}{IEEE Transactions on Automatic Control}}, vol.~62, no.~8, pp. 3861--3876, Aug. 2017, arXiv:1609.06408 [cs, math]. [Online]. Available: \url{http://arxiv.org/abs/1609.06408}
\BIBentrySTDinterwordspacing

\bibitem{berkenkamp_safe_2017}
\BIBentryALTinterwordspacing
F.~Berkenkamp, M.~Turchetta, A.~Schoellig, and A.~Krause, ``Safe {Model}-based {Reinforcement} {Learning} with {Stability} {Guarantees},'' in \emph{Advances in {Neural} {Information} {Processing} {Systems}}, vol.~30.\hskip 1em plus 0.5em minus 0.4em\relax Curran Associates, Inc., 2017. [Online]. Available: \url{https://proceedings.neurips.cc/paper_files/paper/2017/hash/766ebcd59621e305170616ba3d3dac32-Abstract.html}
\BIBentrySTDinterwordspacing

\bibitem{mnih_human-level_2015}
\BIBentryALTinterwordspacing
V.~Mnih, K.~Kavukcuoglu, D.~Silver, A.~A. Rusu, J.~Veness, M.~G. Bellemare, A.~Graves, M.~Riedmiller, A.~K. Fidjeland, G.~Ostrovski, S.~Petersen, C.~Beattie, A.~Sadik, I.~Antonoglou, H.~King, D.~Kumaran, D.~Wierstra, S.~Legg, and D.~Hassabis, ``Human-level control through deep reinforcement learning,'' \emph{Nature}, vol. 518, pp. 529--533, Feb. 2015, aDS Bibcode: 2015Natur.518..529M. [Online]. Available: \url{https://ui.adsabs.harvard.edu/abs/2015Natur.518..529M}
\BIBentrySTDinterwordspacing

\bibitem{mnih_playing_2013}
\BIBentryALTinterwordspacing
V.~Mnih, K.~Kavukcuoglu, D.~Silver, A.~Graves, I.~Antonoglou, D.~Wierstra, and M.~Riedmiller, ``\BIBforeignlanguage{en}{Playing atari with deep reinforcement learning},'' Dec. 2013, arXiv:1312.5602. [Online]. Available: \url{http://arxiv.org/abs/1312.5602}
\BIBentrySTDinterwordspacing

\bibitem{silver_mastering_2017}
\BIBentryALTinterwordspacing
D.~Silver, J.~Schrittwieser, K.~Simonyan, I.~Antonoglou, A.~Huang, A.~Guez, T.~Hubert, L.~Baker, M.~Lai, A.~Bolton, Y.~Chen, T.~Lillicrap, F.~Hui, L.~Sifre, G.~van~den Driessche, T.~Graepel, and D.~Hassabis, ``\BIBforeignlanguage{en}{Mastering the game of {Go} without human knowledge},'' \emph{\BIBforeignlanguage{en}{Nature}}, vol. 550, no. 7676, pp. 354--359, Oct. 2017, publisher: Nature Publishing Group. [Online]. Available: \url{https://www.nature.com/articles/nature24270}
\BIBentrySTDinterwordspacing

\bibitem{nguyen-tuong_local_2008}
\BIBentryALTinterwordspacing
D.~Nguyen-tuong, J.~Peters, and M.~Seeger, ``\BIBforeignlanguage{en}{Local gaussian process regression for real time online model learning},'' in \emph{\BIBforeignlanguage{en}{Advances in {Neural} {Information} {Processing} {Systems}}}, vol.~21.\hskip 1em plus 0.5em minus 0.4em\relax Curran Associates, Inc., 2008. [Online]. Available: \url{https://proceedings.neurips.cc/paper/2008/hash/01161aaa0b6d1345dd8fe4e481144d84-Abstract.html}
\BIBentrySTDinterwordspacing

\bibitem{yang_safe_2022}
\BIBentryALTinterwordspacing
T.-Y. Yang, T.~Zhang, L.~Luu, S.~Ha, J.~Tan, and W.~Yu, ``\BIBforeignlanguage{en}{Safe reinforcement learning for legged locomotion},'' in \emph{\BIBforeignlanguage{en}{2022 {IEEE}/{RSJ} {International} {Conference} on {Intelligent} {Robots} and {Systems} ({IROS})}}, Oct. 2022, pp. 2454--2461, iSSN: 2153-0866. [Online]. Available: \url{https://ieeexplore.ieee.org/document/9982038}
\BIBentrySTDinterwordspacing

\bibitem{pecka_safe_2014}
M.~Pecka and T.~Svoboda, ``\BIBforeignlanguage{en}{Safe exploration techniques for reinforcement learning – an overview},'' in \emph{\BIBforeignlanguage{en}{Modelling and {Simulation} for {Autonomous} {Systems}}}, J.~Hodicky, Ed.\hskip 1em plus 0.5em minus 0.4em\relax Cham: Springer International Publishing, 2014, pp. 357--375.

\bibitem{gu_review_2024}
\BIBentryALTinterwordspacing
S.~Gu, L.~Yang, Y.~Du, G.~Chen, F.~Walter, J.~Wang, and A.~Knoll, ``\BIBforeignlanguage{en}{A {Review} of {Safe} {Reinforcement} {Learning}: {Methods}, {Theory} and {Applications}},'' May 2024, arXiv:2205.10330 [cs]. [Online]. Available: \url{http://arxiv.org/abs/2205.10330}
\BIBentrySTDinterwordspacing

\bibitem{wang_deepsafempc_2024}
\BIBentryALTinterwordspacing
X.~Wang, H.~Pu, H.~J. Kim, and H.~Li, ``\BIBforeignlanguage{en}{{DeepSafeMPC}: deep learning-based model predictive control for safe multi-agent reinforcement learning},'' Mar. 2024, arXiv:2403.06397. [Online]. Available: \url{http://arxiv.org/abs/2403.06397}
\BIBentrySTDinterwordspacing

\bibitem{fisac_general_2019}
\BIBentryALTinterwordspacing
J.~F. Fisac, A.~K. Akametalu, M.~N. Zeilinger, S.~Kaynama, J.~Gillula, and C.~J. Tomlin, ``A {General} {Safety} {Framework} for {Learning}-{Based} {Control} in {Uncertain} {Robotic} {Systems},'' \emph{IEEE Transactions on Automatic Control}, vol.~64, no.~7, pp. 2737--2752, Jul. 2019, conference Name: IEEE Transactions on Automatic Control. [Online]. Available: \url{https://ieeexplore.ieee.org/document/8493361}
\BIBentrySTDinterwordspacing

\bibitem{haarnoja_soft_2018}
\BIBentryALTinterwordspacing
T.~Haarnoja, A.~Zhou, P.~Abbeel, and S.~Levine, ``\BIBforeignlanguage{en}{Soft actor-critic: off-policy maximum entropy deep reinforcement learning with a stochastic actor},'' Aug. 2018, arXiv:1801.01290. [Online]. Available: \url{http://arxiv.org/abs/1801.01290}
\BIBentrySTDinterwordspacing

\bibitem{galloway_torque_2015}
\BIBentryALTinterwordspacing
K.~Galloway, K.~Sreenath, A.~D. Ames, and J.~W. Grizzle, ``\BIBforeignlanguage{en}{Torque saturation in bipedal robotic walking through control lyapunov function-based quadratic programs},'' \emph{\BIBforeignlanguage{en}{IEEE Access}}, vol.~3, pp. 323--332, 2015, conference Name: IEEE Access. [Online]. Available: \url{https://ieeexplore.ieee.org/document/7079382}
\BIBentrySTDinterwordspacing

\bibitem{mehra_adaptive_2015}
\BIBentryALTinterwordspacing
A.~Mehra, W.-L. Ma, F.~Berg, P.~Tabuada, J.~W. Grizzle, and A.~D. Ames, ``\BIBforeignlanguage{en}{Adaptive cruise control: experimental validation of advanced controllers on scale-model cars},'' \emph{\BIBforeignlanguage{en}{2015 American Control Conference (ACC)}}, pp. 1411--1418, Jul. 2015, conference Name: 2015 American Control Conference (ACC) ISBN: 9781479986842 Place: Chicago, IL, USA Publisher: IEEE. [Online]. Available: \url{http://ieeexplore.ieee.org/document/7170931/}
\BIBentrySTDinterwordspacing

\bibitem{ames_rapidly_2014}
\BIBentryALTinterwordspacing
A.~D. Ames, K.~Galloway, K.~Sreenath, and J.~W. Grizzle, ``\BIBforeignlanguage{en}{Rapidly exponentially stabilizing control lyapunov functions and hybrid zero dynamics},'' \emph{\BIBforeignlanguage{en}{IEEE Transactions on Automatic Control}}, vol.~59, no.~4, pp. 876--891, Apr. 2014, conference Name: IEEE Transactions on Automatic Control. [Online]. Available: \url{https://ieeexplore.ieee.org/document/6709752}
\BIBentrySTDinterwordspacing

\bibitem{vamvoudakis_online_2010}
\BIBentryALTinterwordspacing
K.~G. Vamvoudakis and F.~L. Lewis, ``\BIBforeignlanguage{en}{Online actor–critic algorithm to solve the continuous-time infinite horizon optimal control problem},'' \emph{\BIBforeignlanguage{en}{Automatica}}, vol.~46, no.~5, pp. 878--888, May 2010. [Online]. Available: \url{https://linkinghub.elsevier.com/retrieve/pii/S0005109810000907}
\BIBentrySTDinterwordspacing

\end{thebibliography}

\begin{IEEEbiography}[{\includegraphics[width=1in,height=1.25in,clip,keepaspectratio]{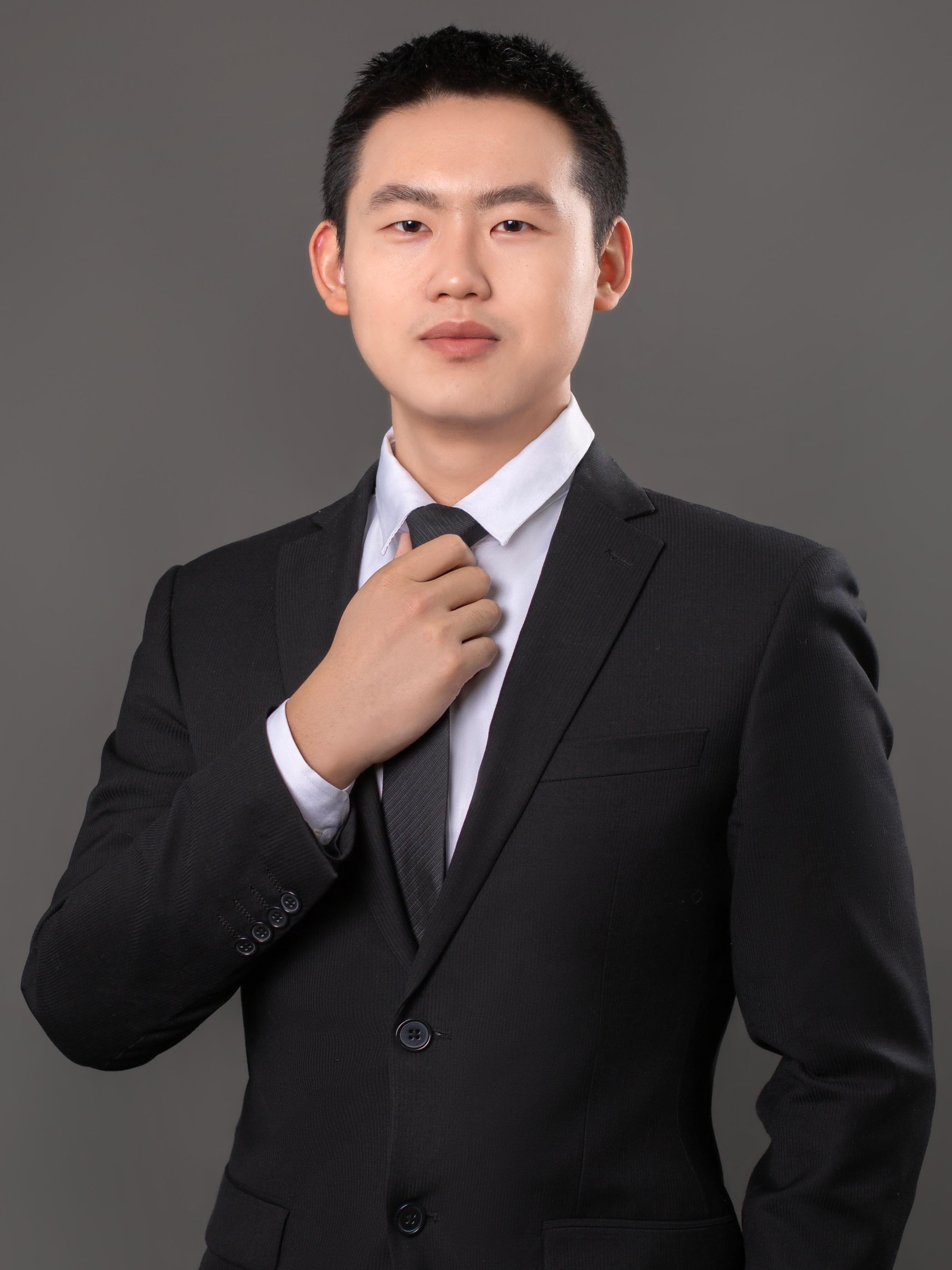}}]{Donghe Chen} was born in Nanping, China, in 2001. He received his B.E. degree in Beihang University (a.k.a. Beijing University of Aeronautics and Astronautics), Beijing, China in 2023. He is now a postgraduate in the School of Astronautics, Beihang University, China.
  His current interests include dynamic system identification, deep reinforcement learning and real-time optimal control.
\end{IEEEbiography}

\begin{IEEEbiography}[{\includegraphics[width=1in,height=1.25in,clip,keepaspectratio]{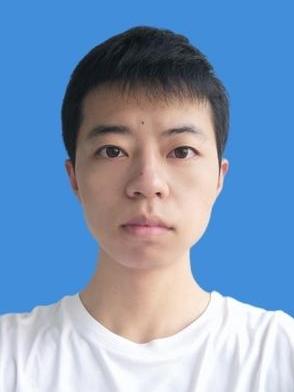}}]{Han Wang} received the B.S. degree in aerospace engineering from Sun Yat-Sen University in 2022. He is now a postgraduate in the School of Astronautics, Beihang University, China.
  His current research interests include terminal guidance, optimal control, and intelligent guidance and control.
\end{IEEEbiography}

\begin{IEEEbiography}[{\includegraphics[width=1in,height=1.25in,clip,keepaspectratio]{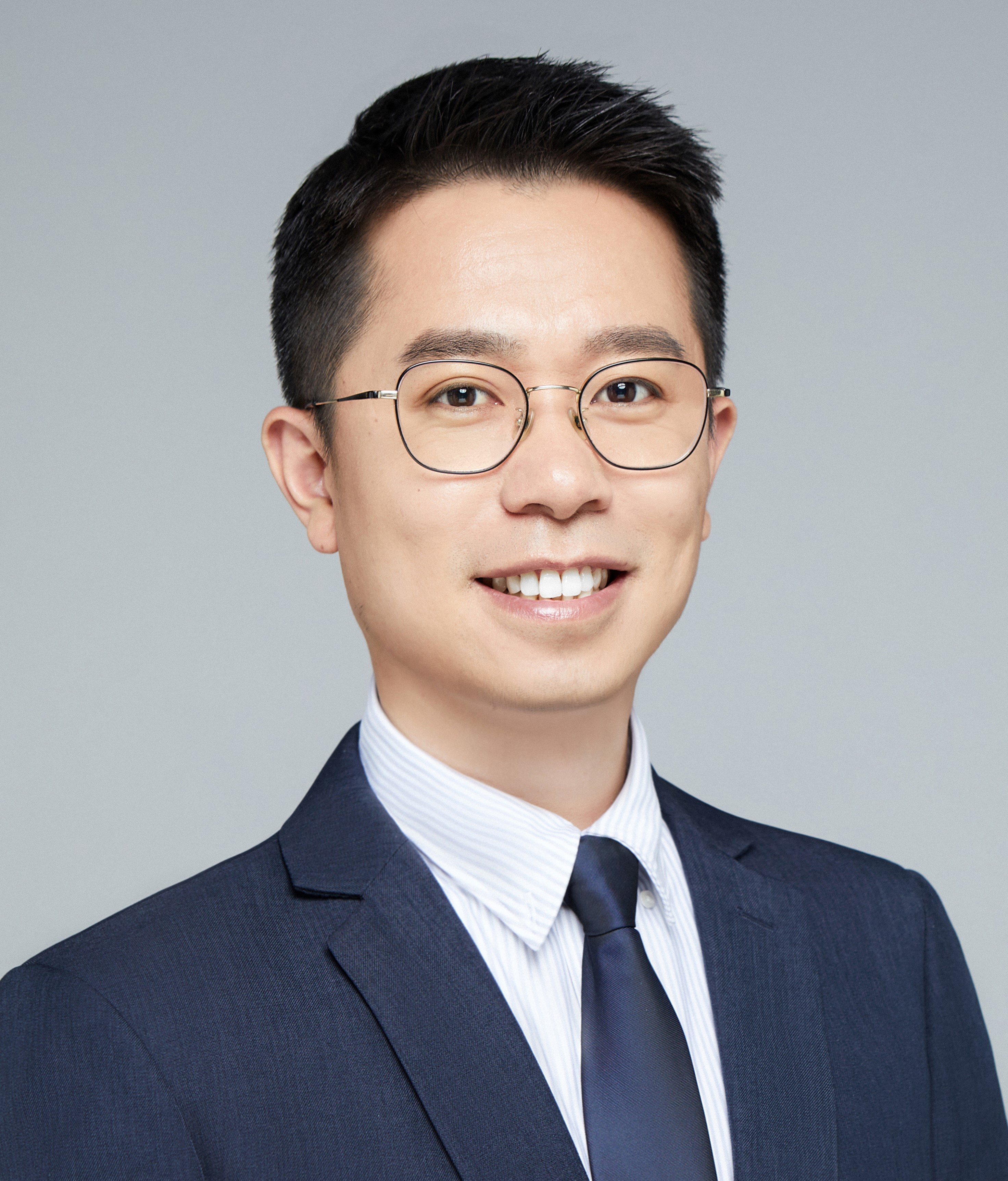}}]{Lin Cheng} received his Ph.D. degree in  Automation Science and Electrical Engineering from Beihang University (a.k.a. Beijing University of Aeronautics and Astronautics), Beijing, China in 2017, and is working as an associate professor in School of Astronautics, Beihang University, Beijing, China.

  His current interests include guidance and control, trajectory optimization, deep reinforcement learning and real-time optimal control with AI.
\end{IEEEbiography}

\begin{IEEEbiography}[{\includegraphics[width=1in,height=1.25in,clip,keepaspectratio]{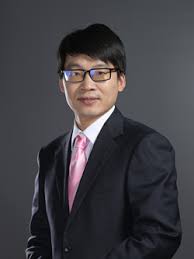}}]{Shengping Gong} received the B.S. degree in aerospace engineering from the National University of Defense
  Technology, Changsha, China, in 2004, and the Ph.D. degree in astrodynamics and control from Tsinghua
  University, Beijing, China, in 2008.

  After spending a year as a Postdoctoral Researcher, he became an Assistant Researcher and then an
  Associate Professor with the School of Aerospace Engineering, Tsinghua University. In 2021, he was a
  Professor with the School of Astronautics, Beihang University. His research interests include the dynamics
  and control of spacecraft, trajectory optimization, solar sail, and celestial mechanics
\end{IEEEbiography}

\end{document}